\documentclass[pdflatex,sn-mathphys-num]{sn-jnl}

\usepackage{graphicx}%
\usepackage{multirow}%
\usepackage{amsmath,amssymb,amsfonts}%
\usepackage{amsthm}%
\usepackage{mathrsfs}%
\usepackage[title]{appendix}%
\usepackage{xcolor}%
\usepackage{textcomp}%
\usepackage{manyfoot}%
\usepackage{booktabs}%
\usepackage{algorithm}%
\usepackage{algorithmicx}%
\usepackage{algpseudocode}%
\usepackage{listings}%
\usepackage{stfloats}
\usepackage{subfig}

\theoremstyle{thmstyleone}%
\newtheorem{theorem}{Theorem}
\theoremstyle{thmstyletwo}%

\theoremstyle{thmstylethree}%

\begin{document}

\title[Article Title]{SILSM: A Sustainable Interactive Level Set Method for Progressive Refinement}


\author[1]{\fnm{Jiachen} \sur{Song}}\email{25B312016@stu.hit.edu.cn}
\author[1]{\fnm{Dazhi } \sur{Zhang}}\email{zhangdazhi@hit.edu.cn}
\author*[1]{\fnm{Fanghui} \sur{Song}}\email{fanghuisong@stu.hit.edu.cn}

\author[1]{\fnm{Zhichang} \sur{Guo}}\email{mathgzc@hit.edu.cn}
\author[1]{\fnm{Shengzhu} \sur{Shi}}\email{mathssz@hit.edu.cn}

\affil*[1]{School of Mathematics, Harbin Institute of Technology, Harbin 150006, China}


\abstract{Interactive segmentation aims to precisely isolate target objects using sparse user guidance. However, traditional methods often suffer from heavy interaction burdens and parameter sensitivity, while deep learning approaches struggle with data dependency and iterative instability. Motivated by these limitations, we propose the Sustainable Interactive Level Set Method (SILSM). The proposed level set evolution equation incorporates interaction, regularization, and segmentation terms. Specifically, high-order regularization is employed to maintain numerical stability, and unlike traditional methods, we decouple user guidance into an independent interaction term to enable direct manual control over the zero-level set evolution. Furthermore, we develop a numerical algorithm tailored for multiple interactions, which facilitates dynamic refinement by effectively updating the segmentation results based on sequential user inputs. We theoretically demonstrate that the high-order term provides stronger regularization constraints than the conventional length term, while the interaction term ensures segmentation strictly within the user-selected region. Experimental results further demonstrate that the proposed method is robust to interactive inputs, achieves competitive performance at the first interaction, and supports stable multi-round interactions with progressively improved segmentation quality.}




\maketitle

\section{Introduction}
Image segmentation is a fundamental task in computer vision \cite{yu2023techniques,ramesh2021review,sun2025efficient}, aiming to partition an image into semantically distinct regions at the pixel level. By accurately delineating object boundaries and spatial extents, it enables a fine-grained decomposition of visual content, thereby serving as a core enabling technology for numerous downstream applications such as medical image analysis \cite{bogensperger2025flowsdf,chen2025uncertainty}, autonomous driving perception \cite{feng2020deep,vobecky2025unsupervised}, industrial inspection \cite{zhang2025small,liu2023survey} and other areas \cite{allabadi2025learning,jain2023oneformer}.

Selective segmentation, unlike global approaches that aim to partition the entire image, focuses exclusively on extracting a specific object of interest defined by the user. This task-specific paradigm is essential for complex scenes where irrelevant background structures must be ignored. To achieve this, interactive strategies have been widely developed, generally falling into two categories: discrete graph-based optimization and continuous variational frameworks. The first category includes methods such as Graph Cuts \cite{rother2004grabcut}, Random Walks \cite{grady2006random}, and Geodesic approaches \cite{bai2007geodesic}. While successful in many scenarios, these discrete methods often rely on minimal-length constraints, which can over-smooth boundaries and struggle to capture intricate shapes. Consequently, there remains a critical need for continuous models that can better respect local geometry and user intent.

In contrast to discrete approaches, the second category leverages continuous variational frameworks—specifically Level Set Methods (LSMs) \cite{osher1988fronts}—to impose geometric constraints on the desired object via user-placed markers. Gout et al. \cite{gout2005segmentation} pioneered this direction by defining a distance function from a set of markers and incorporating it into the edge-based Geometric Active Contour (GAC) model \cite{caselles1997geodesic}. However, this approach inherited the GAC's sensitivity to noise and boundary leakage at weak edges. To address these limitations, subsequent research shifted towards integrating regional statistics. For instance, Badshah et al. \cite{badshah2010image} incorporated the region-based Chan-Vese (CV) model \cite{chan2001active} to enhance robustness against noise and blurred boundaries. Building on this, Rada and Chen \cite{rada2012new} introduced a dual Level Set Function (LSF) to simultaneously handle global segmentation and selective focusing. Further advancements include the introduction of area constraint terms by Rada et al. \cite{rada2013improved} and generalized average fitting terms by Ali et al. \cite{ali2021image} for multi-region tasks. Additionally, Spencer et al. \cite{spencer2015convex} extended convex relaxation techniques to interactive segmentation, while Rada et al. \cite{rada2018selective} proposed specific formulations for intensity inhomogeneous images. Common to all these variational approaches is the core strategy of embedding geometric constraints into global segmentation models to achieve user-guided, selective results.

Despite the mathematical rigor of these approaches, directly embedding geometric constraints into the variational framework introduces critical limitations. A primary drawback is that the segmentation performance becomes hypersensitive to the precise location of user inputs. Since the geometric constraint term rigidly forces the contour towards specific markers, even slight deviations in marker placement can lead to suboptimal results, thereby imposing a high burden on the user to provide accurate initializations near object boundaries. Furthermore, balancing this geometric term against image-based terms introduces additional parameters that often require extensive manual tuning. Most critically, these methods typically operate under a static, one-shot paradigm. They process the initial user input as a fixed constraint without a mechanism to integrate sequential corrections. This lack of iterative refinement contradicts the fundamental nature of human-computer interaction, where users expect to progressively guide the segmentation process.

To address these limitations, several neural network-based interactive segmentation methods have recently been proposed \cite{wong2024scribbleprompt,kirillov2023segment,luo2021mideepseg,liu2023simpleclick}. While achieving impressive performance, these approaches, however, suffer from two major drawbacks. First, they require extensive annotated datasets, including both segmentation masks and interaction annotations, which entails substantial human effort. More critically, they often produce unstable or inconsistent segmentation results across multiple interaction rounds, frequently deviating from user expectations. This instability stems largely from the inherent lack of theoretical guarantees and limited interpretability of their underlying network mechanisms.

Motivated by the limitations of existing approaches such as the prohibitive data dependency and instability of deep learning models and the inability of traditional variational frameworks to integrate sequential user corrections, we propose the \emph{Sustainable Interactive Level Set Method} (\textbf{SILSM}). We address the sensitivity and fixity issues of prior geometric constraint models by decoupling the interaction mechanism from the segmentation process. This novel formulation transforms the static interaction paradigm into a dynamic one, allowing for continuous, stable user refinement. Our main contributions can be summarized as follows:

\begin{itemize}
	\item[1)] We formulate interaction information as an independent first-order term, strictly decoupling it from the inherent segmentation energy functional. This structural independence ensures that user interventions act as precise guidance without compromising the model's intrinsic ability to capture complex object boundaries. Unlike traditional coupled approaches, our design prevents interaction constraints from overwhelming image features, thereby maintaining high segmentation accuracy while effectively eliminating the need for complex, case-specific parameter tuning.
	\item[2)] We incorporate a fourth-order regularization term to govern the evolution dynamics of the level set function. Unlike traditional lower-order regularization, this higher-order constraint effectively suppresses numerical oscillations and spurious noise. This ensures the robust evolution of the segmentation contour, maintaining high geometric quality and stability throughout the iterative process.
	\item[3)] Within the level set framework, we establish a progressive refinement mechanism that supports continuous user interaction. This design enables the model to accumulate user guidance over successive rounds, facilitating sequential multi-object segmentation, fine-grained structure extraction, and targeted error correction. Consequently, the segmentation quality steadily improves as additional constraints are integrated.
\end{itemize}

The remainder of this paper is organized as follows: Section \ref{sec2} reviews several interactive segmentation models closely related to our work. Section \ref{sec3} presents our proposed model and the corresponding algorithm, followed by a brief overview of several representative tasks that the model can accomplish. Section \ref{sec4} demonstrates that our model is robust to variations in interactive input, supports multiple interaction rounds, and maintains stable performance throughout the iterative refinement process, as evidenced by experiments on both medical image segmentation and real-world image segmentation. Finally, conclusions of this work are discussed in Section \ref{sec5}.

\section{Related work}\label{sec2}
In this section, we briefly review the related works and mathematical formulations that inspired the development of our framework. We begin by defining the standard notations used in this domain.

Let $\Omega \subset \mathbb{R}^2$ denote the image domain and $\Gamma$ represent an evolving closed curve within it. The evolution of $\Gamma$ is described by $\phi(x,y,t)$, where $t$ denotes an artificial time parameter. The curve $\Gamma(t)$ at time $t$ is given by the zero level set $\{(x,y) \mid \phi(x,y,t) = 0\}$, with $\phi > 0$ defining its interior and $\phi < 0$ its exterior. Finally, $I(x,y)$ represents the grayscale intensity of the original image at pixel $(x,y)$.

\subsection{CV model \cite{chan2001active}}
The CV model \cite{chan2001active} is a region-based active contour model derived from a simplification of the Mumford-Shah energy minimization problem \cite{mumford1989optimal}. It was proposed to overcome the limitations of traditional edge-based active contour models, which rely on image gradients and often fail on images with weak or missing boundaries, or significant noise.

The core idea of the CV model is to leverage regional intensity homogeneity rather than edge gradients. It assumes the image comprises two approximately piecewise-constant regions. The model evolves an initial contour by minimizing an energy functional that balances the fidelity of regional intensity fitting, the length of the contour for regularization, and the area inside the contour. The energy functional $E_{CV}(c_1, c_2, \Gamma)$ is defined as follows:
$$
\begin{aligned}
	E_{CV}\left(c_1, c_2, \Gamma \right)= & \mu \cdot \operatorname{Length}(\Gamma)+\nu \cdot \operatorname{Area}(\operatorname{inside}(\Gamma)) \\
	& +\lambda_1 \int_{\operatorname{inside}(\Gamma)}\left|I(x, y)-c_1\right|^2 \mathrm{d}x\mathrm{d}y \\
	& +\lambda_2 \int_{\operatorname{outside}(\Gamma)}\left|I(x, y)-c_2\right|^2 \mathrm{d}x\mathrm{d}y,
\end{aligned}
$$
where $c_1$ and $c_2$ represent the average intensities inside and outside the contour $\Gamma$, respectively. The nonnegative parameters $\mu$, $\nu$, $\lambda_1$, and $\lambda_2$ control the trade-off between contour smoothness, area constraint, and regional fitting fidelity.

The CV model is considered a milestone in region-based active contour segmentation. By providing a simplified yet effective formulation of the Mumford-Shah model within a level set framework, it overcomes the dependency on image gradients and offers a more robust and general-purpose segmentation framework, particularly for images with weak boundaries or noise.

\subsection{Interactive segmentation via level set method}

\subsubsection{Gout model \cite{gout2005segmentation}}

Let $S=\{(x_i,y_i)\in\Omega, \, 1\le i \le n\}$ be a set of $n $ marker points near the boundary of the target object. Based on these interaction points, the distance function $ d(x,y)$ can be constructed. A common choice is the Euclidean distance to the nearest marker:
$$
d(x, y)=\min _{\left(x_i, y_i\right) \in S}\left|(x, y)-\left(x_i, y_i\right)\right|,
$$
Alternatively, a Gaussian-based multiplicative form can be used \cite{le2008geodesic}
$$
d(x, y)=\prod_{i=1}^n\left(1-e^{-\frac{\left(x-x_i\right)^2}{2 \sigma^2}} e^{-\frac{\left(y-y_i\right)^2}{2 \sigma^2}}\right) \qquad \forall(x, y) \in \Omega.
$$
The scale parameter $\sigma$ adjusts the spatial influence of each marker. Notably, $d(x,y) \to 0$ near the markers and approaches $1$ far from them. Gout et al. proposed to extract the object boundary $\Gamma$ by minimizing the following energy functional:
$$
E(\Gamma) = \int_\Gamma d(x,y)g(|\nabla I(x,y)|) \mathrm{d}s,
$$
where
$$
g(|\nabla I(x,y)|) = \frac{1}{1+|\nabla I(x,y)|^2}.
$$
Here, $g$ acts as an edge-stopping function that approaches zero at high-gradient boundaries. The evolution of the curve is guided by the model towards areas where both $d$ and $g$ exhibit low values. However, accurate segmentation of target objects becomes challenging when images contain strong noise or exhibit blurred boundaries.

\subsubsection{Badshah model \cite{badshah2010image}}
To improve segmentation robustness on noisy images, Badshah et al. introduced an improved Gout model by integrating the CV model. This method significantly enhances the capability to capture objects with weak or blurred boundaries and effectively suppresses high-frequency noise, leading to the following energy:
$$
\begin{aligned}
	E(\Gamma) = &\mu \int_\Gamma d(x,y)g(|\nabla I|)\mathrm{d}s + \lambda_1 \int_{\text{inside}(\Gamma)}|I(x,y)-c_1|^2\mathrm{d}x\mathrm{d}y \\&+ \lambda_2 \int_{\text{inside}(\Gamma)}|I(x,y)-c_2|^2\mathrm{d}x\mathrm{d}y.
\end{aligned}
$$
A critical issue arises when solving the model via standard time marching: the regional terms can dominate to the extent that the solution degenerates into the piecewise constant result of the original CV model. This causes a complete loss of interactive capability, resulting in an undesired global segmentation. Moreover, the model introduces additional parameters and remains highly sensitive to the initial placement of user markers.

\subsubsection{Zhao model \cite{zhao2022new}}
Zhao et al. \cite{zhao2022new} proposed a two-stage approach. In the initial stage, a structure-preserving smoothing model is applied to the input image $I$ by solving:
$$
\begin{aligned}
	\min _{u \in W^{1,2}(\Omega)} E(u):= &\frac{1}{2} \int_{\Omega}\left(1-g\left(\left|\nabla_{\mathrm{GSG}} f(x, y)\right|\right)\right)(u-I)^2(x, y) \mathrm{d}x\mathrm{d}y \\
	& + \frac{\alpha}{2} \int_{\Omega} g\left(\left|\nabla_{\mathrm{GSG}} f(x, y)\right|\right)|\nabla u|^2(x, y) \mathrm{d}x\mathrm{d}y,
\end{aligned}
$$
where $\nabla_{\text{GSG}}$ denotes the global sparse gradient operator \cite{zhang2017global}.
This model represents an alternative paradigm in interactive segmentation, where user interaction is leveraged to perform selective smoothing on the original image. This process effectively mitigates the interference of complex background textures on the final segmentation result. The first term serves as a weighted data fidelity term, while the second term acts as an edge-aware regularization that preserves structural information during smoothing.
In the second stage, an improved Gout model is applied to the smoothed image $u$ to detect the target boundary. The marker points are used only to initialize the active contour in this stage, rather than being directly involved in the subsequent evolution.

However, this method presents notable limitations. First, the smoothing process inevitably alters the original image fidelity, risking the loss of fine structural details. Second, the two-stage architecture increases parameter complexity, imposing a heavy burden for manual tuning. Finally, by restricting markers solely to the initialization phase, the model lacks continuous geometric constraints, making the segmentation results hypersensitive to the initial user input.

\subsection{Curve evolution and level set method}

The general equation of curve evolution can be simplified as:
\begin{equation}\label{curve evolution}
	\frac{\partial C}{\partial t} = \beta \mathbf{N},
\end{equation}
where $C$ denotes the evolving curve, $\beta$ is normal velocity, and $\mathbf{N}$ is the unit normal vector. However, this explicit equation struggles to handle topological changes of curves, such as merging and splitting.

To address this issue, Osher and Sethian proposed the LSM in 1988 \cite{osher1988fronts}. Its core idea is to implicitly embed the low-dimensional evolving curve into a high-dimensional function $\phi(x,y,t)$, where the evolving curve is represented as the zero-level set of $\phi$, i.e., 
$$C(t) = \{(x,y) | \phi(x,y,t) = 0\}.$$ 
Using the implicit function theorem and the chain rule, the evolution of the curve is transformed into a Hamilton Jacobi-type partial differential equation (PDE) for $\phi$:
$$
\frac{\partial \phi}{\partial t} + \beta |\nabla \phi| = 0.
$$
This method converts curve evolution into the evolution of a high-dimensional scalar field, facilitating the use of mature numerical methods for solution. It can naturally handle topological changes without additional logical judgments and is easily extensible to the evolution problems of 3D and higher-dimensional surfaces. The Eulerian formulation elegantly converts curve evolution into the evolution of a fixed grid-based scalar field. Consequently, it can naturally accommodate complex topological changes without any special logic, benefits from a wealth of mature numerical techniques for PDEs, and is readily extensible to surfaces in three or higher dimensions.

\subsection{The MBE regularization \cite{song2024re}}
To address common issues in level set evolution (e.g., the need for periodic reinitialization and sensitivity to initialization), Song et al. \cite{song2024re,song2025} proposed a high-order level set variational segmentation method incorporating regularization derived from Molecular Beam Epitaxy (MBE) equations with slope selection. This equation models the epitaxial growth of thin films and can be obtained from the following energy functional:
$$
E_\text{MBE} (\phi)= \int_\Omega \frac{\alpha}{2} |\Delta \phi|^2 + \frac{1}{4}(|\nabla \phi|^2 - 1)^2 \mathrm{d} x.
$$

The proposed energy functional incorporates two physically-motivated regularization terms to ensure numerical stability and segmentation accuracy. The biharmonic term, $\frac{\alpha}{2} |\Delta \phi|^2$, imposes higher-order smoothness, effectively suppressing high-frequency noise and irregularities. Complementing this, the double-well potential term, $\frac{1}{4}(|\nabla \phi|^2 - 1)^2$, penalizes deviations from the signed distance property ($|\nabla \phi| = 1$), thereby preserving the structural integrity of the level set function without requiring costly re-initialization. Together, these terms reshape the energy landscape to favor stable, topologically coherent evolution, significantly reducing sensitivity to initialization while robustly capturing complex geometries.

\section{Methodology}\label{sec3}

\subsection{Motivation and framework overview}
Addressing the limitations of static initialization and parameter sensitivity in prior works, we introduce a framework designed for continuous, iterative refinement. The key innovation lies in treating the segmentation process not as a single minimization problem, but as a synergistic combination of data fidelity and user intent. In this paper, our model integrates energy-driven optimization with geometry-driven evolution, formulated as follows:

\begin{equation}\label{SILSM equation}
    \begin{aligned}
        \frac{\partial \phi}{\partial t} = & \underbrace{- \delta (\phi) (\lambda_1 (I-c_1)^2 - \lambda_2 (I-c_2)^2)}_{\text{Segmentation Term}} +
        \underbrace{ F|\nabla \phi|}_{\text{Interaction Term}} \\
        & +\underbrace{\mu\left(-\alpha \Delta^2 \phi + \text{div}\left((|\nabla \phi|^2-1)\nabla \phi\right)\right)}_{\text{Regularization Term}}
    \end{aligned}
\end{equation}

As shown in Eq. \ref{SILSM equation}, the evolution is governed by three synergistic components. The energy-guided variational segmentation term exploits global intensity statistics to drive the contour towards object boundaries, while the energy-guided variational regularization term ensures numerical stability and boundary smoothness through intrinsic geometric constraints. Complementing these, the geometry-guided interactive information prior term $F|\nabla \phi|$ translates user inputs into an external driving force. Uniquely, this formulation achieves a decoupling of interaction from segmentation: rather than embedding user constraints as rigid penalties within the energy functional, we treat interaction as an independent, additive velocity field, preserving the model's variational stability while enabling flexible, continuous refinement. In the following subsections, we will provide a detailed derivation and analysis of each component.

\subsubsection{Geometry-guided interactive information prior term}
To establish the interaction term, we draw upon the fundamental relationship between curve evolution and level set propagation. Let $C(t)$ denote the evolving curve and $\mathbf{N}$ be its outward unit normal. The geometric evolution equation is given by 
$$
\frac{\partial C}{\partial t} = F \mathbf{N},
$$ 
which corresponds to the level set equation 
\begin{equation}\label{lsf evolution}
	\frac{\partial \phi}{\partial t} = F|\nabla \phi|.
\end{equation}
Here, we adopt the convention that $\phi > 0$ represents the object interior $\Omega_{\text{in}}$ and $\phi < 0$ represents the exterior $\Omega_{\text{out}}$.

In classical curve shortening flow, the speed function is typically defined as the mean curvature, i.e., $F = \text{div}(\nabla \phi / |\nabla \phi|)$. While this formulation effectively smooths boundaries by penalizing length, we will demonstrate in subsequent sections that relying solely on curvature is insufficient for responsive user-guided segmentation, as it lacks directional specificity towards user-designated targets.

To ensure that the contour strictly follows user instructions, we design $F$ as a piecewise constant function. Specifically, to drive the zero level set to contract from the background towards the object boundary, we define $F$ as follows:
\begin{equation}
F(x,y) =\begin{cases}
-a, & \text{outside} \, \Omega_{\text{interested}}\\
0, & \text{inside} \,\Omega_{\text{interested}} 
\end{cases}
\end{equation}
where $\Omega_{\text{interested}}$ is the area that users are interested in and $a >0$ is a constant speed parameter. Under this definition, for points situated in the exterior of the region of interest, the negative speed $F = -a$ acts in opposition to the outward unit normal direction. Geometrically, this generates a constant inward force, compelling the zero level set to shrink continuously until it encounters the true object boundary or is counterbalanced by the segmentation term. From the perspective of the level set evolution, this negative speed causes the value of $\phi$ to decrease monotonically over time. Consequently, these points eventually transition to negative values, thereby being explicitly classified as part of the background $\Omega_{\text{out}}$. This formulation effectively translates the geometric prior of "background suppression" into a variational contraction force.

It is crucial to highlight a fundamental theoretical distinction regarding the coupling mechanism in this framework compared to traditional models. In standard variational approaches, the evolution equation is strictly derived as the gradient descent flow of a unified scalar energy functional i.e., $\frac{\partial \phi}{\partial t} = -\frac{\delta E}{\delta \phi}$. This inherently couples user constraints with image-based terms, forcing them to compete within the same minimization process. In contrast, the interaction term proposed here is inherently non-variational; it cannot be integrated back into a static energy potential form. Unlike traditional methods that embed geometric constraints as soft penalty terms—often leading to a conflict between data fidelity and user intent—our approach injects user guidance directly as an extrinsic kinematic force. This formulation achieves a structural decoupling of interaction from segmentation: by bypassing the energy minimization constraints, the user input serves as a direct velocity command rather than a passive energy penalty. This ensures that user instructions are prioritized and executed independently, without being compromised by the intrinsic resistance of the variational framework.

\subsubsection{Energy-guided regularization term}

For traditional variational models, the length regularization term $\int_\Omega \delta(\phi(x))|\nabla \phi(x)| \text{dx}$ is commonly employed to enforce boundary smoothness. A well-known limitation of this first-order formulation is that the level set function $\phi$ tends to become excessively flat or steep during evolution, deviating from the signed distance property. Such deviation leads to an invalid curvature computation $ \text{div}(\nabla \phi / |\nabla \phi|) $, thereby necessitating periodic reinitialization, which is computationally expensive and may introduce additional numerical errors.

To overcome this issue, we adopt the MBE regularization term \cite{song2025}. This term imposes a stronger and more physically motivated constraint on the entire level set function, rather than only its zero level set. The corresponding energy functional is given by
\begin{equation}\label{MBE}
	E_{MBE} = \int_\Omega \frac{\alpha}{2} |\Delta \phi|^2 + \frac{1}{4}(|\nabla \phi|^2 - 1)^2 \mathrm{d}x.
\end{equation}
The corresponding gradient descent flow is
\begin{equation}\label{MBE evolution}
	\frac{\partial \phi}{\partial t} = -\alpha \Delta^2 \phi + \text{div}\left((|\nabla \phi|^2 -1)\nabla \phi\right).
\end{equation}
Compared to length term, the MBE regularization offers following advantages: 1) The biharmonic term $\alpha \Delta^2 \phi$ ensures higher-order smoothness; 2) The nonlinear term drives $|\nabla \phi|$ towards 1, naturally maintaining $\phi$ as an approximate signed distance function and eliminating the need for reinitialization; 3) The resulting evolution equation is free from the singular term $1/|\nabla \phi|$, leading to enhanced numerical stability.

\subsubsection{Energy-guided segmentation term}

For clarity and as a representative baseline, we adopt the region-based segmentation term from the classic CV model \cite{chan2001active}. The fitting term of CV model formulates image segmentation as an energy minimization problem, aiming to find an optimal contour implicitly represented by the zero-level set of a LSF $\phi$ that separates the image into foreground and background regions. Its energy functional is defined as:
\begin{align}\label{eq:fitting term}
	E_{\text{seg}}(\phi,c_1,c_2) = & \lambda_1 \int_\Omega H_\epsilon (\phi) (I(x,y) - c_1)^2  \mathrm{d}x\mathrm{d}y \nonumber\\
	& + \lambda_2 \int_\Omega (1 - H_\epsilon(\phi))(I(x,y)-c_2)^2  \mathrm{d}x\mathrm{d}y,
\end{align}
where $c_1$ and $c_2$ are the average intensity values of the foreground and background regions separated by the zero-level contour of $\phi$, defined as:
$$
c_1 = \frac{\int_\Omega H_\epsilon(\phi) I(x,y)\mathrm{d}x\mathrm{d}y}{\int_\Omega H_\epsilon(\phi)\mathrm{d}x\mathrm{d}y},\; c_2 = \frac{\int_\Omega(1-H_\epsilon(\phi)) I(x,y)\mathrm{d}x\mathrm{d}y}{\int_\Omega (1-H_\epsilon(\phi))\mathrm{d}x\mathrm{d}y},
$$
$H_\epsilon(\cdot)$ is smooth approximations of Heaviside function,
$$
H_\epsilon(\phi) = \frac{1}{2} \left(1+\frac{2}{\pi}\text{arctan}(\frac{\phi}{\epsilon})\right),
$$
with $\epsilon>0$ being a small smoothing parameter.

Minimizing \eqref{eq:fitting term} via calculus of variations yields the following gradient descent flow
\begin{equation}\label{CV evolution}
	\frac{\partial \phi}{\partial t} = -\delta_\epsilon(\phi)\left(\lambda_1(I-c_1)^2-\lambda_2(I-c_2)^2\right),
\end{equation}
where $\delta_\epsilon(\cdot)$ is smooth approximations of the Dirac delta function $\delta(\cdot)$ given by
$$
\delta_\epsilon(\phi) = \frac{1}{\pi} \frac{\epsilon}{\epsilon^2+\phi^2}.
$$

It is worth noting that while we adopt the CV model as a representative baseline for region-based segmentation, the proposed SILSM framework is inherently modular. Thanks to the decoupling of interaction from segmentation, this term can be flexibly substituted with other variational functionals—such as edge-based models \cite{caselles1997geodesic} or local intensity fitting models \cite{li2008minimization}—to adapt to diverse image characteristics without altering the core interactive mechanism.

In summary, the proposed SILSM framework effectively combines geometry-driven interaction with energy-driven optimization. By integrating a sustainable, multi-round interactive mechanism into the level set method, our approach ensures both flexibility and mathematical stability. This design aims to achieve segmentation accuracy comparable to state-of-the-art deep learning models while offering superior interpretability and theoretical convergence guarantees, all without the need for extensive training data.

\subsection{Model analysis}
In this section, we provide a theoretical and empirical analysis of the proposed SILSM to validate its robustness and effectiveness. We first present a theorem to rigorously characterize the kinematic behavior of the geometry-guided interaction term, proving that it guarantees the contraction of the zero level set in the exterior of the region of interest. Subsequently, we demonstrate that the MBE regularization imposes a stronger, higher-order geometric constraint compared to the traditional length term. Finally, through a experiment, we verify the indispensability of coupling the interaction, regularization, and segmentation terms for stable contour evolution.

\subsubsection{Convergence analysis of the interactive term}
In this subsection, we establish the convergence property of the geometry-guided interaction term defined in Eq. \eqref{lsf evolution}. We specifically analyze the kinematic behavior of the level set function in the exterior of the region of interest. According to our design, the speed function in this region is a constant negative value $F = -a$ ($a > 0$). Consequently, the evolution equation reduces to a fundamental Hamilton-Jacobi equation governing constant inward motion.The following theorem serves as the theoretical guarantee for our "background suppression" mechanism. By setting the propagation speed $c = -a$, we rigorously demonstrate that the zero-level set undergoes monotonic contraction. This ensures that any contour initialized or evolving outside the target region will inevitably shrink and vanish in finite time, thereby enforcing the user's geometric constraints.

\begin{theorem}[Zero-Level Set Contraction]\label{thm:contraction}
	Let $\phi(\mathbf{x},t): \Omega \times [0, T) \to \mathbb{R}$ satisfy the level set evolution equation
	\begin{equation}\label{eq:thm_evolution}
		\frac{\partial \phi}{\partial t} = c |\nabla \phi|,
	\end{equation}
	where $c < 0$ is a constant. Assume the initial LSF $\phi_0(\mathbf{x}) \in C^2(\Omega)$ is chosen such that its zero level set $\Gamma_0$ is a compact, smooth hypersurface enclosing an interior region $\Omega_{\text{in}}$ (where $\phi_0 > 0$), and $\Omega_{\text{out}}$ (where $\phi_0 < 0$) represents the exterior.
	
	Let $\Gamma(t) := \{\mathbf{x} \in \Omega \mid \phi(\mathbf{x}, t) = 0\}$ denote the evolving zero level set. Then the following properties hold:
	\begin{enumerate}
		\item \textbf{Inward Motion:} The interface $\Gamma(t)$ moves strictly inward (towards $\Omega_{\text{in}}$) with a constant normal speed $|c|$.
		
		\item \textbf{Finite-Time Collapse:} The zero level set will shrink and vanish in finite time. Specifically, there exists a time $T^* > 0$ such that $\Gamma(t) = \emptyset$ for all $t > T^*$. An upper bound for the collapse time is
		\[
		T^* \leq \frac{R_{\mathrm{in}}}{|c|},
		\]
		where $R_{\mathrm{in}}$ is the inradius of the initial region $\Omega_{\text{in}}$ (the radius of the largest inscribed ball).
	\end{enumerate}
\end{theorem}

\begin{proof}
	\paragraph{Step 1: Geometric interpretation of the motion}
	Since $\phi_0 \in C^2$ and the speed is constant, standard theory for Hamilton-Jacobi equations guarantees the existence of a viscosity solution. We analyze the motion geometrically. The normal velocity $V_n$ of the level set $\Gamma(t)$ is given by the standard level set relationship:
	\[
	V_n = -\frac{1}{|\nabla \phi|} \frac{\partial \phi}{\partial t}.
	\]
	Substituting the evolution equation \eqref{eq:thm_evolution} into this expression yields:
	\[
	V_n = -\frac{c |\nabla \phi|}{|\nabla \phi|} = -c.
	\]
	We now consider the direction. The unit normal vector is defined as $\mathbf{n} = \frac{\nabla \phi}{|\nabla \phi|}$. Since we adhere to the convention that $\phi > 0$ inside the object and $\phi < 0$ outside, the gradient $\nabla \phi$ points from the exterior to the interior. Thus, $\mathbf{n}$ is the inward-pointing normal.
	
	The velocity vector of the contour is given by $\vec{V} = V_n \mathbf{n} = -c \mathbf{n}$. Since $c < 0$, the scalar term $-c$ is positive. This means the velocity vector points in the same direction as $\mathbf{n}$, which is the \textbf{inward} direction. Consequently, every point on the contour moves inward with a constant speed $|c|$, causing the enclosed region $\Omega_{\text{in}}$ to shrink monotonically.
	
	\paragraph{Step 2: Upper bound for the collapse time}
	To estimate the collapse time, we employ the comparison principle. Consider a comparison function $\psi(\mathbf{x},t)$ whose zero level set is a sphere evolving with the same normal speed.
	
	Let $B(\mathbf{x}_c, R_{\text{in}})$ be the largest ball of radius $R_{\text{in}}$ centered at $\mathbf{x}_c$ that is fully contained within the initial interior $\Omega_{\text{in}}$. The evolution of this sphere's radius $R(t)$ under the constant inward speed $|c|$ is governed by:
	\[
	\frac{\mathrm{d}R}{\mathrm{d}t} = -|c|, \quad R(0) = R_{\text{in}}.
	\]
	The solution is $R(t) = R_{\text{in}} - |c|t$. The sphere collapses to a point when $R(t) = 0$, which occurs at time $T_{\text{sphere}} = \frac{R_{\text{in}}}{|c|}$.
	
	Since the initial sphere is contained within $\Gamma_0$, and both interfaces evolve inward with the same speed, the comparison principle dictates that the sphere must remain contained within $\Gamma(t)$ as long as $\Gamma(t)$ exists. Therefore, $\Gamma(t)$ cannot survive longer than the sphere. The interface $\Gamma(t)$ must collapse at a time $T^* \leq T_{\text{sphere}}$.
\end{proof}

In summary, Theorem \ref{thm:contraction} provides the theoretical guarantee for the proposed background suppression mechanism. It confirms that by setting a negative speed parameter ($c<0$) in the exterior region, the zero level set is forced to contract monotonically. This kinematic property ensures that the contour effectively escapes background clutter and is driven inevitably towards the target object boundary.

\subsubsection{Complete substitution of length regularization}
While the interaction term derived in Theorem \ref{thm:contraction} effectively drives the zero-level set toward the object boundary, the introduction of such external kinematic forces can compromise numerical stability. Specifically, the constant pushing force $F$ may introduce local irregularities or sharp discontinuities in the level set function, particularly in the presence of noise. Traditional length-based regularization often fails to counteract these higher-order oscillations, leading to a loss of the signed distance property. To ensure stable evolution, our framework incorporates the MBE regularization term \eqref{MBE}.

The following theorem theoretically establishes that the MBE term imposes a strictly stronger geometric constraint than the traditional length term. By bounding the length energy with the Laplacian energy, we guarantee that minimizing the MBE functional implicitly controls boundary smoothness with higher-order precision.

\begin{theorem}\label{th1}
    Let $\Omega \subset \mathbb{R}^n$ be a bounded domain with a boundary of class $C^{1,1}$. Let $\phi \in H^2 (\Omega) \cap H_0^1 (\Omega)$ be a function such that $\phi \neq 0$ in the interior of $\Omega$. Assume the boundary condition $|\nabla \phi| \geq m > 0$ holds on $\partial \Omega$. Let $\delta$ denote the Dirac delta function. There exists a positive constant $C$ dependent only on the region $\Omega$ and $m$ such that the following inequality holds
    \begin{equation}\label{inequality}
        \int_\Omega \delta(\phi) |\nabla \phi| \, \mathrm{d}x \leq C \int_\Omega |\Delta \phi|^2 \,\mathrm{d}x.
    \end{equation}
\end{theorem}

\begin{proof}
    Let $\mathbf{F} = \phi \nabla \phi$. By the divergence theorem and the identity $\nabla \cdot (\phi \nabla \phi) = |\nabla \phi|^2 + \phi \Delta \phi$ we have
    $$
    \int_\Omega |\nabla \phi|^2 \, \mathrm{d}x + \int_\Omega \phi \Delta \phi \, \mathrm{d}x = \int_{\partial \Omega}\phi \nabla \phi \cdot \mathbf{n} \, \mathrm{d}s.
    $$
    Since $\phi$ vanishes on the boundary $\partial \Omega$, the right side equals 0. Thus
    $$
    \int_\Omega |\nabla \phi|^2 \,\mathrm{d}x = -\int_\Omega \phi \Delta \phi \, \mathrm{d}x.
    $$
    The Cauchy-Schwarz inequality implies
    $$
    \int_\Omega |\nabla \phi|^2 \, \mathrm{d}x \leq \|\phi\|_{L^2} \|\Delta \phi\|_{L^2}.
    $$
    The Poincar{\'e} inequality $\|\phi\|_{L^2} \leq C_P\|\nabla \phi\|_{L^2}$ allows us to write
    $$
    \|\nabla \phi\|_{L^2}^2 \leq C_P \|\nabla \phi\|_{L^2} \|\Delta \phi\|_{L^2}.
    $$
    Dividing by $\|\nabla \phi\|_{L^2}$ yields
    \begin{equation}\label{1}
        \| \nabla \phi \|_{L^2} \leq C_P \| \Delta \phi \|_{L^2}.
    \end{equation}
    
    We now invoke the Trace Theorem and Elliptic Regularity\cite{evans2022partial}. Since the domain has a $C^{1,1}$ boundary the $H^2$ norm is controlled by the Laplacian
    $$
    \|\phi\|_{H^2} \leq C_E \|\Delta \phi\|_{L^2}.
    $$
    The Trace Theorem states that the boundary norm of the gradient is controlled by the global $H^2$ norm
    $$
    \|\nabla \phi\|_{L^2(\partial \Omega)} \leq C_T \|\phi\|_{H^2}.
    $$
    Combining these two estimates gives
    \begin{equation}\label{2}
        \int_{\partial \Omega}|\nabla \phi|^2 \, \mathrm{d}s \leq K \|\Delta \phi\|_{L^2}^2,
    \end{equation}
    where $K$ is a combined constant.
    
    Using the coarea formula interpretation of the Dirac delta function and the assumption that $\phi \neq 0$ inside $\Omega$, the integral on the left side becomes
    $$
    \int_\Omega \delta(\phi) |\nabla \phi| \, \mathrm{d}x = \int_{\partial \Omega} 1 \, \mathrm{d}s.
    $$
    Using the condition $|\nabla \phi| \geq m$ on the boundary
    $$
    \int_{\partial \Omega} 1 \, \mathrm{d}s \leq \int_{\partial \Omega} \frac{|\nabla \phi|^2}{m^2} \,\mathrm{d}s.
    $$
    Substituting inequality \eqref{2} into this expression results in
    \begin{align*}
        \int_\Omega \delta(\phi) |\nabla \phi| \, \mathrm{d}x &\leq \frac{K}{m^2} \| \Delta \phi \|^2_{L^2} \\
        &= C \int_\Omega |\Delta \phi|^2 \,\mathrm{d}x.
    \end{align*}
\end{proof}

The $L^2$ gradient flow of Eq. \eqref{MBE} is
\begin{equation*}\label{MBEflow}
	\left\{\begin{aligned}
		&\frac{\partial \phi}{\partial t}=-\alpha \Delta^2 \phi+\operatorname{div}\left(\left(|\nabla \phi|^2-1\right) \nabla \phi\right), &&(x, t) \in \Omega \times(0, T], \\
		&\phi(x, 0)=\phi_0(x), && x \in \Omega.
	\end{aligned}\right.
\end{equation*}

From the perspective of function space and geometric flow theory, the method represents a fundamental paradigm shift: it transitions from traditional $BV$-type geometric regularization---which relies on first-order derivatives to minimize boundary length---to an $H^2$-type morphological regularization based on a fourth-order energy, designed to control the variation of interface curvature. Within the MBE energy functional, the biharmonic term $|\Delta \phi|^2$ acts as a surface diffusion current, imposing a strong smoothing constraint, while the nonlinear term $(|\nabla \phi|^2-1)^2$ serves as a non-equilibrium diffusion term, driving the gradient magnitude of the level set function $\phi$ toward unity. This mechanism inherently preserves the signed distance property during evolution, thereby eliminating the need for reinitialization.

This shift in the regularization mechanism fundamentally alters the objective of the model. Rather than solely seeking the shortest boundary, the high-order terms implicitly regularize the segmentation contour by controlling its curvature, leading to the evolution of a structurally consistent interface. In the sense of $\Gamma$-convergence, the limiting functional of such high-order energies does not degenerate into the traditional perimeter functional \cite{Attouch2014}. This theoretically circumvents the inherent problems of interface collapse and loss of fine details associated with models based on mean curvature flow. The resulting structural difference provides a rigorous theoretical foundation for preserving sharp corners, thin elongated structures, and multi-scale geometric features in complex images.

\subsubsection{Integration and necessity of model components}
\begin{figure}[!ht]
    \centering
    \includegraphics[width=2.5in]{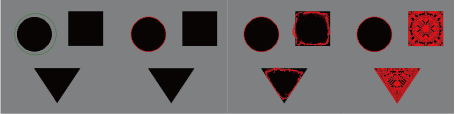}
    \caption{Comparison of segmentation performance. Panel 1: The green circle indicates the region of interest. Panel 2: Proposed method. Panel 3: Curvature-driven interaction. Panel 4: No regularization. The process ran for $300$ iterations.}
    \label{fig1}
\end{figure}

Having analyzed the interaction and regularization terms individually, we now demonstrate that their integration with the segmentation term is essential for a robust interactive framework. The proposed method relies on the synergy of three forces: the segmentation term drives data fidelity, the constant-speed interaction term enforces user constraints against background clutter, and the MBE regularization maintains the structural integrity of the LSF. The complete evolution equation of our SILSM is formulated as:
\begin{equation}\label{the evolution equation}
	\begin{aligned}
		\frac{\partial \phi}{\partial t} = &- \delta (\phi) (\lambda_1 (I-c_1)^2 - \lambda_2 (I-c_2)^2) +\mu(-\alpha \Delta^2 \phi \\&+ \text{div}((|\nabla \phi|^2-1)\nabla \phi)) + F|\nabla \phi|,
	\end{aligned}
\end{equation}
where $c_1$ and $c_2$ represent the mean intensities of the foreground and the background within the region of interest, respectively:
$$
c_1 = \frac{\int_\Omega H_\epsilon(\phi) I(x)\mathrm{d}x}{\int_\Omega H_\epsilon(\phi)\mathrm{d}x},\; c_2 = \frac{\int_{\Omega_\text{interested}}(1-H_\epsilon(\phi)) I(x)\mathrm{d}x}{\int_{\Omega_\text{interested}} (1-H_\epsilon(\phi))\mathrm{d}x}.
$$
Here, $\Omega_\text{interested}$ denotes the user-specified region of interest. To validate the necessity of each component, we conducted an ablation study using a synthetic setup with segmentation weights uniformly set to $\lambda_1 = \lambda_2 = 1$. The interaction force $F$ is applied exclusively outside the region of interest (i.e., $F=0$ inside $\Omega_\text{interested}$), and each simulation runs for $300$ iterations. We compare three configurations:
\begin{enumerate}
\item \textbf{Proposed Method} $(\mu = 1, \alpha = 5, F=-500)$: Utilizes our constant inward force with MBE regularization.
\item \textbf{Curvature-Driven Interaction} $(\mu = 1, \alpha = 5, F=10^{20} \kappa)$: Replaces the constant force with a mean curvature flow ($\kappa = \text{div}(\nabla \phi/|\nabla \phi|)$).
\item \textbf{No Regularization} $(\mu = 0, F=-500)$: Utilizes the constant force but removes the MBE regularization.
\end{enumerate}
The segmentation results are presented in Fig. \ref{fig1}. The first panel delineates the region of interest with a green circle.

\begin{figure}[!ht]
    \centering
    \includegraphics[width=4.5in]{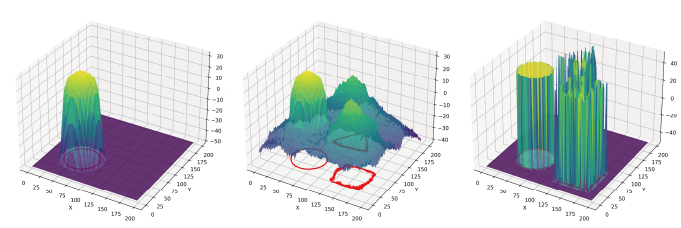}
    \caption{Final LSFs resulting from the three segmentation processes. Left: The proposed method. Middle: Curvature-Driven interaction. Right: No regularization.}
    \label{fig2}
\end{figure}

The experimental results highlight the critical role of each term. As observed in Fig. \ref{fig1}, the curvature-driven flow (Panel 3) fails to drive the contour completely into the region of interest despite an extremely large coefficient. This confirms that curvature flow, being geometry-dependent, is insufficient to overcome local image gradients or geometric resistance, whereas our constant force (Panel 2) acts as a deterministic kinematic driver that guarantees monotonic contraction. Furthermore, the necessity of the MBE regularization is evident when examining the final LSFs in Fig. \ref{fig2}. While the non-regularized case (Panel 4 in Fig. \ref{fig1}) appears to segment the object, its corresponding LSF (Fig. \ref{fig2}, Right) develops severe irregularities and steep gradients, deviating from the signed distance property. In contrast, the proposed method (Fig. \ref{fig2}, Left) maintains a smooth and numerically stable LSF. Thus, the synergy of all three components is indispensable: the constant interaction force ensures convergence into the region of interest, while the MBE regularization ensures the stability required for the segmentation term to accurately capture the object boundary.

\subsection{Multi-Stage interaction via direct LSF reconstitution}

To address the challenge of local minima inherent in non-convex energy functionals, we introduce a multi-stage interactive strategy. While traditional methods often attempt to segment an image in a single pass, they frequently fail to capture nested or complex structures due to energy barriers. Our framework overcomes this by treating segmentation as a progressive process: the user can intervene at any stage to reconstitute the LSF, effectively resetting the evolution state to target finer details that were initially missed.

This Direct Assignment mechanism modifies the LSF directly according to user input. Unlike passive regularization terms, this intentional perturbation acts as a strong, task-specific prior, enabling the evolution to escape undesirable local minima and ``jump'' across energy barriers. This capability is fundamental to our multi-stage paradigm, as it allows the user to guide the model from a coarse global segmentation to a refined local result.

We demonstrate this mechanism on a synthetically generated image of $50\times50$ pixels. The image consists of an almost white background (intensity 240), a central black square of size $30\times30$ (intensity 0), and a gray circle with a radius of 10 (intensity 128) located at the center of the square. The segmentation is performed using the CV model.

\begin{figure}[!ht]
	\centering
	\includegraphics[width=3.5in]{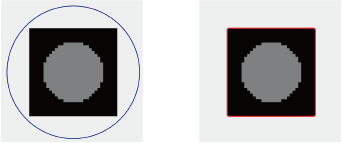}
	\caption{Left: The blue contour indicates the initial boundary. Right: The red contour shows the final segmentation result.}
	\label{fig3}
\end{figure}

The initial contour and the final segmentation result are shown in Fig. \ref{fig3}. It has been empirically demonstrated that piecewise constant functions serve as more efficient initializations for LSMs compared to signed distance functions. For our experimental framework, we adopt the following function as the initial LSF to maintain equilibrium between the internal and external regions:
\begin{equation}\label{initial level set function}
	\phi^0(x, y)=\left\{\begin{array}{ll}
		+c, & \text { if } (x, y) \in \Omega_{\text{in}} \\
		-1, & \text { if } (x, y) \in \Omega_{\text{out}}.
	\end{array}\right.
\end{equation}
Here, the parameter $c$ is set to $\frac{\text{Area}_{\text{out}}}{\text{Area}_{\text{in}}}$ to satisfy the condition $\int_\Omega \phi^0(x,y) \mathrm{dxdy} = 0$.

To analyze the capture of the central circle, we evaluate the CV energy functional by defining the interior of a centered circle with radius $r$ ($0 \le r \le 15$) as the object exterior. The energy is calculated as:
\begin{equation}
	\begin{aligned}
		E_{CV} &= \int_\Omega H(\phi)(I-c_1)^2\mathrm{d}x + \int_\Omega(1-H(\phi))(I-c_2)^2\mathrm{d}x \\
		&=\left\{
		\begin{aligned}
			&(0-c_1)^2(900-100\pi )+ (128-c_1)^2(100\pi-\pi r^2) \\
			& +(128-c_2)^2 \pi r^2 + (240-c_2)^2 1600,\quad 0\le r \le 10, \\[2ex]
			&0+(128-c'_2)^2 100 \pi+(0-c'_2)^2(r^2-100)\pi \\
			& +(240-c'_2)^2 1600, \quad 10 < r \le 15,
		\end{aligned}
		\right.
	\end{aligned}
\end{equation}
where the intensity means are given by:
\begin{align*}
	c_1=\frac{128\pi(100-r^2)}{900-\pi r^2}, \quad c_2 = \frac{128\pi r^2+384000}{\pi r^2+1600}, \\ c'_1=0, \quad c'_2=\frac{12800\pi + 384000}{\pi r^2+1600}.
\end{align*}

\begin{figure}[!ht]
	\centering
	\includegraphics[width=0.28\textwidth]{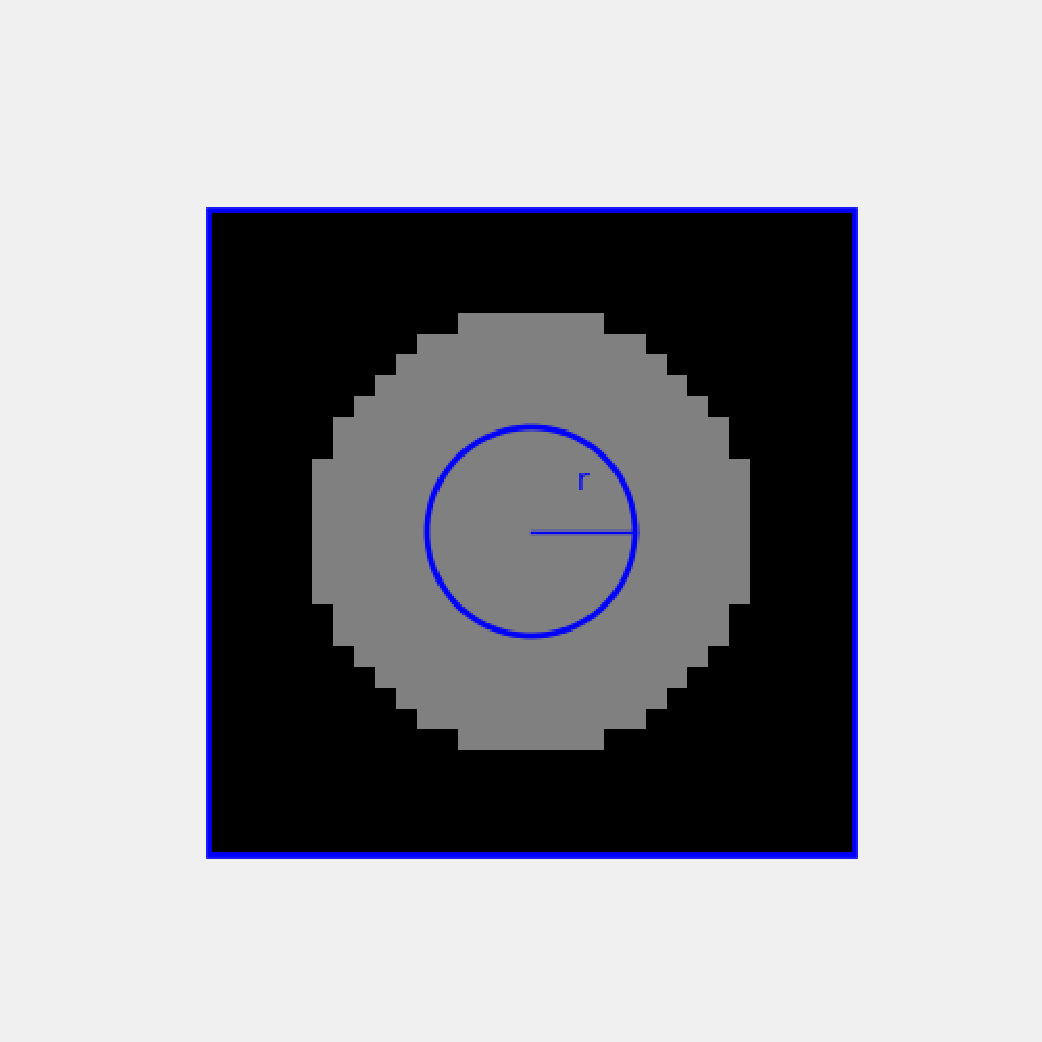}
	\hspace{0.5cm}
	\includegraphics[width=0.4\textwidth]{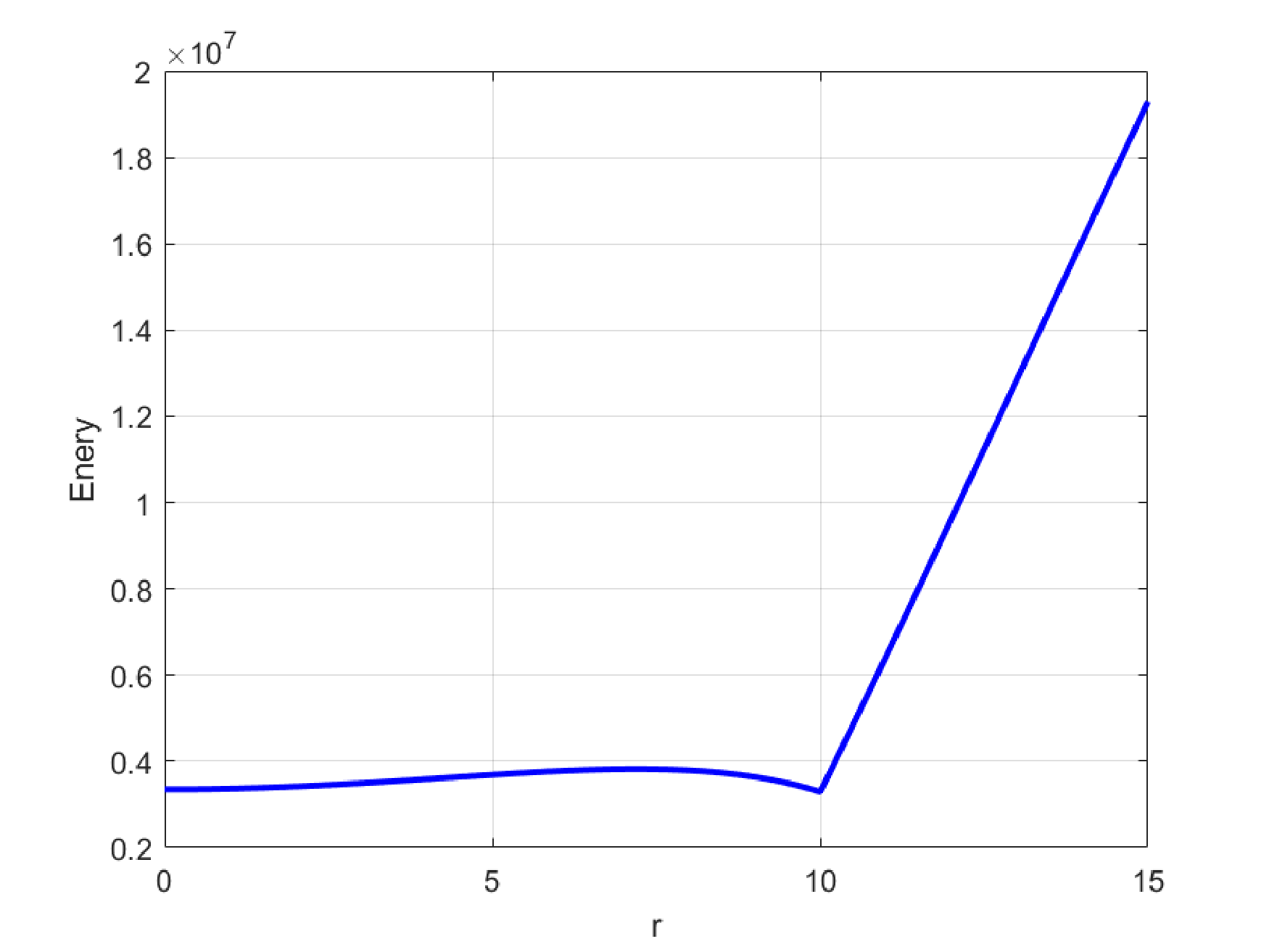}
	\caption{Left: Schematic where the central circular region of radius $r$ is converted to the exterior. Right: Energy profile of the CV model for $0 \le r \le 15$.}
	\label{fig4}
\end{figure}

The energy profile of the CV model is plotted in Fig. \ref{fig4}. The energy $E_{CV}$ is calculated to be $3350480$ for $r=0$ and $3294030$ for $r=10$. This energy gap confirms that the segmentation result in Fig. \ref{fig3} corresponds to the LSF being trapped in a local minimum. To achieve further segmentation, it is necessary to guide the LSF out of this state. 

\begin{figure}[!t]
	\centering
	\includegraphics[width=0.2\textwidth]{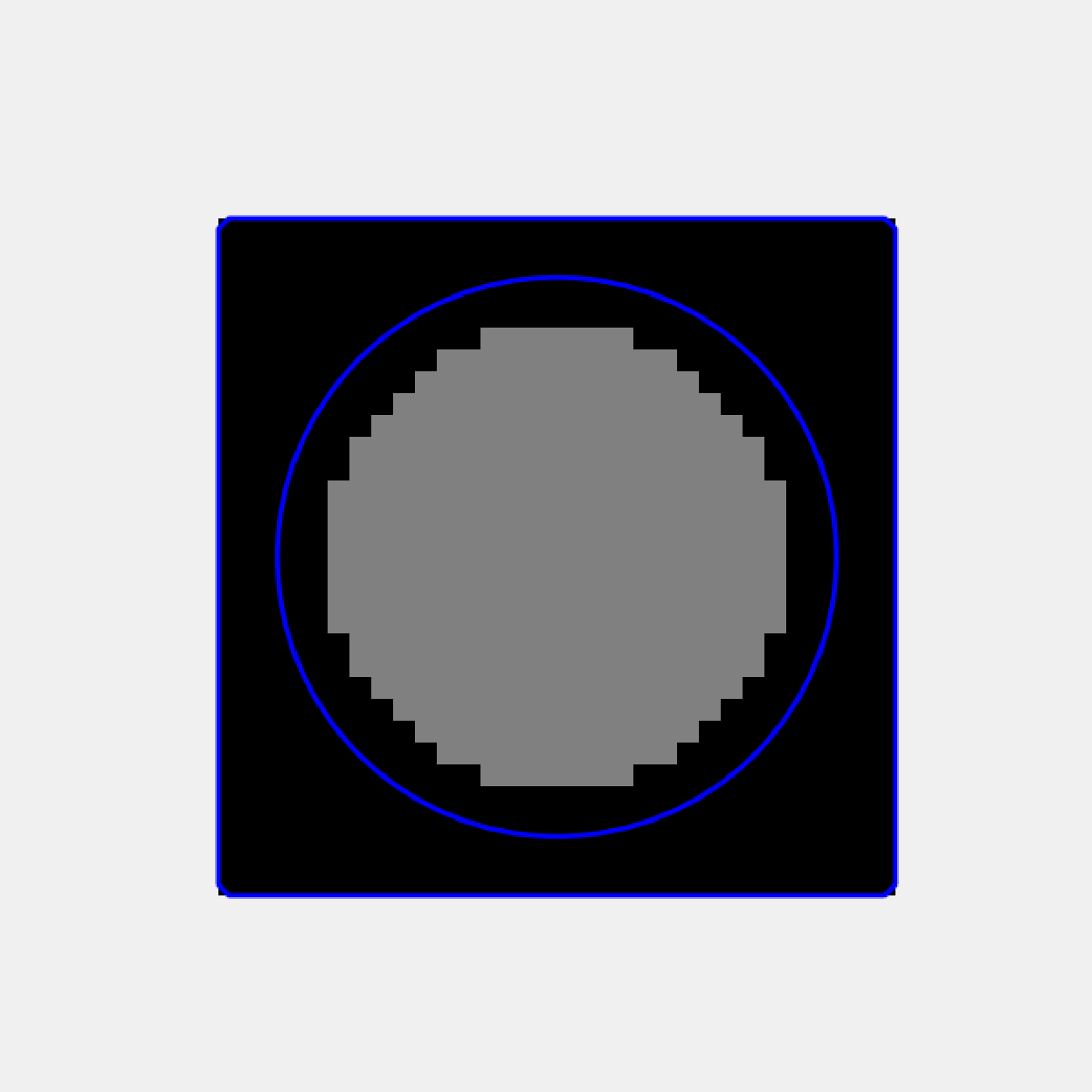}
	\includegraphics[width=0.27\textwidth]{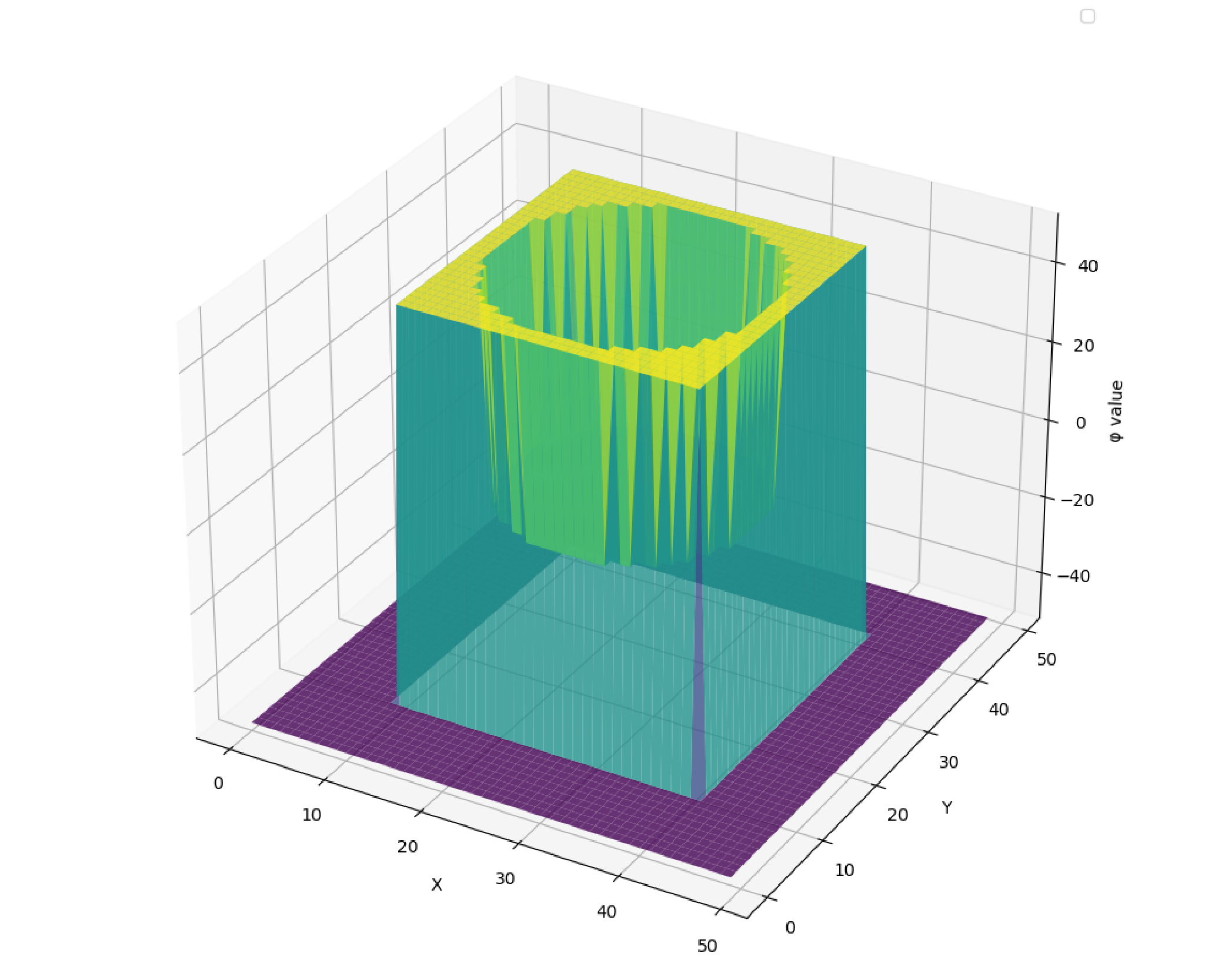}
	\includegraphics[width=0.2\textwidth]{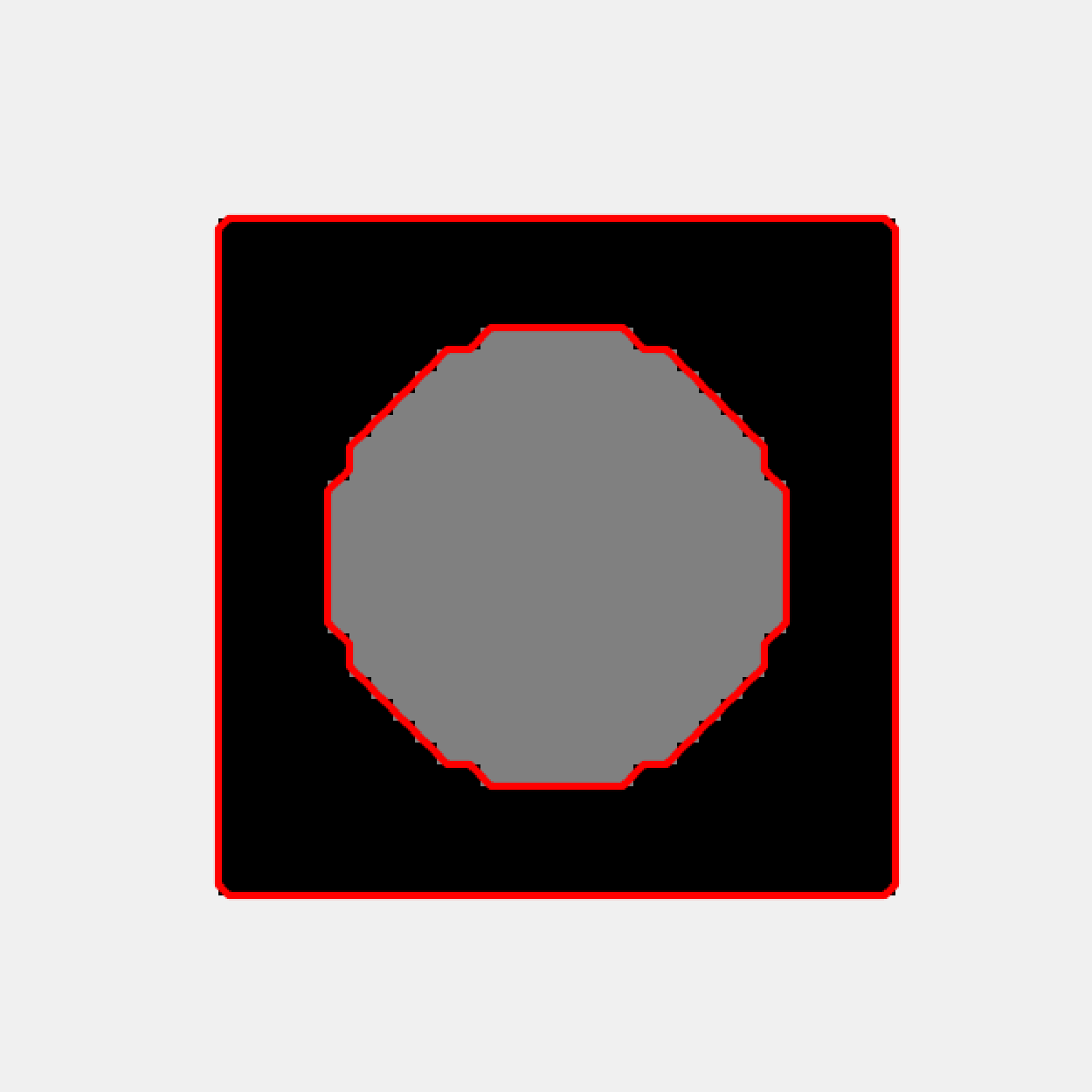}
	\includegraphics[width=0.27\textwidth]{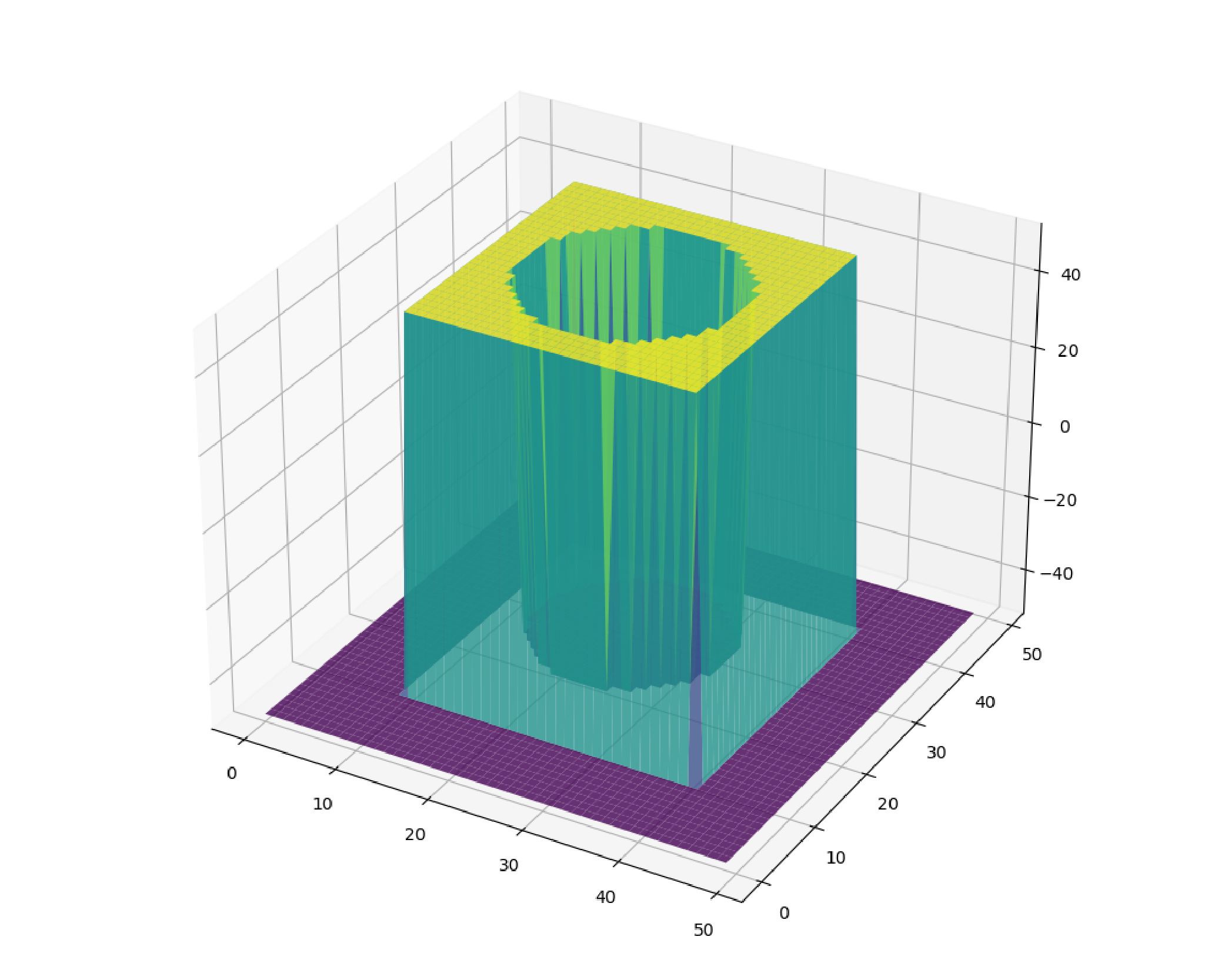}
	\caption{Multi-stage refinement: The LSF from the previous segmentation step is reconstituted and used as the initial state for the subsequent segmentation.}
	\label{fig5}
\end{figure}

We employ a reconstitution strategy to achieve this. Briefly, we select a region; if it was previously interior ($\phi > 0$), its value is reset to a negative constant $-k$, and conversely, if it was exterior ($\phi < 0$), it is set to $+k$. Following the result in Fig. \ref{fig3}, we proceed with further segmentation by assigning a value of $-k$ to the LSF within the central circular region, where $k=\frac{\text{Area}_{\text{out}}}{\text{Area}_{\text{in}}}$ (here, "in" and "out" refer to the interior and exterior of the newly specified region). The initial contour, the intermediate result, and the corresponding LSF are shown in Fig. \ref{fig5}. The final result demonstrates that the central circular region is successfully extracted, validating this approach for segmenting fine details in a multi-stage process.

\subsection{Numerical implementation and algorithm}
The partial differential equation in Eq. \eqref{the evolution equation} is solved numerically on a discrete grid $\Omega_h = \{(x_i, y_j) \mid 1 \le i \le M, 1 \le j \le N\}$ with grid spacing $h=1$. We employ the finite difference method to approximate the spatial derivatives using central difference schemes. The time evolution is handled via a forward Euler scheme, yielding the iterative update rule:
\begin{equation}\label{iteration}
	\phi_{i,j}^{n+1} = \phi_{i,j}^n + \Delta t \cdot \mathcal{L}(\phi_{i,j}^n), \quad n=1,2,...,N
\end{equation}
where $\mathcal{L}(\phi_{i,j}^n)$ represents the discrete approximation of the right-hand side of Eq. \eqref{the evolution equation}.

The spatial derivatives are discretized as follows. Let $D_x$ and $D_y$ denote the central difference operators for the first-order partial derivatives:
\begin{equation*}
	D_x \phi_{i,j} = \frac{\phi_{i+1,j} - \phi_{i-1,j}}{2h}, \quad D_y \phi_{i,j} = \frac{\phi_{i,j+1} - \phi_{i,j-1}}{2h}.
\end{equation*}
Consequently, the gradient magnitude $|\nabla \phi|$ at grid point $(i, j)$ is computed as:
\begin{equation*}
	|\nabla \phi|_{i,j} = \sqrt{(D_x \phi_{i,j})^2 + (D_y \phi_{i,j})^2}.
\end{equation*}
The Laplacian operator $\Delta \phi$ is approximated using the standard five-point stencil:
\begin{equation*}
	\Delta \phi_{i,j} = \frac{\phi_{i+1,j} + \phi_{i-1,j} + \phi_{i,j+1} + \phi_{i,j-1} - 4\phi_{i,j}}{h^2}.
\end{equation*}
For the higher-order MBE regularization term, the biharmonic operator $\Delta^2 \phi$ is computed by applying the discrete Laplacian operator iteratively: $\Delta^2 \phi_{i,j} = \Delta(\Delta \phi_{i,j})$. The nonlinear divergence term is discretized using central differences on the flux components.

Finally, the complete discrete evolution operator $\mathcal{L}(\phi_{i,j})$ is assembled as:
\begin{equation*}
	\begin{aligned}
		\mathcal{L}(\phi_{i,j}) = & -\delta_\epsilon(\phi_{i,j})\left( \lambda_1(I_{i,j}-c_1)^2 - \lambda_2(I_{i,j}-c_2)^2 \right) \\
		& + F_{i,j} |\nabla \phi|_{i,j} + \mu \left( -\alpha \Delta^2 \phi_{i,j} + \text{div}\left((|\nabla \phi|^2-1)\nabla \phi\right)_{i,j} \right).
	\end{aligned}
\end{equation*}
The complete numerical procedure is summarized in Algorithm \ref{alg1}.

\begin{algorithm}[!ht]
	\caption{The Pseudo Code of SILSM}\label{alg1}
	
	\textbf{In the $i$-th user interaction} \\[0.2em]
	
	\textbf{If $i=1$} \\
	\hspace*{0.5cm} \textbf{Input:} The user-defined region of interest $\Omega_1$ and the speed parameter $a_1 > 0$. \\
	\hspace*{0.5cm} \textbf{Initialization:} The regularization coefficient $\mu$, the coefficient $\alpha$ for the fourth-order MBE term, the time step $\Delta t$, the maximum iteration count $N$, the initial LSF from Eq. \eqref{initial level set function}, and areas of interest to users $\Omega_\text{interested}=\emptyset$. \\
	\hspace*{0.5cm} \textbf{Update areas of interest to users:} $\Omega_\text{interested}=\Omega_1$. \\
	\hspace*{0.5cm} \textbf{Construct the velocity field} $F_1$:
	\[
	F_1(x) = \begin{cases}
		-a_1, & x \notin \Omega_1  \\
		0, & x \in \Omega_1 
	\end{cases}
	\]
	\hspace*{0.5cm} \textbf{Repeat $N$ times:} Update the LSF according to Eq. \eqref{iteration}. \\
	\hspace*{0.5cm} \textbf{Output:} The final LSF $\phi^N$ and image with segmentation results. \\
	\textbf{End If} \\[0.3em]
	
	\textbf{If $i > 1$} \\
	\hspace*{0.5cm} \textbf{Input:} \\
	\hspace*{1cm} 1. A single point $P_i$ to determine whether the intended modification area corresponds to the foreground or the background. \\
	\hspace*{1cm} 2. The region $\Omega_i$ of the segmentation result that the user wishes to modify. \\
	\hspace*{1cm} 3. The speed parameter $a_i \ge 0$. \\
	\hspace*{0.5cm} \textbf{Update areas of interest to users:} $\Omega_\text{interested} = \Omega_\text{interested} \cup \Omega_i$. \\
	\hspace*{0.5cm} \textbf{Reconstitution:} Reconstitute $\phi^{(i-1)N}$ as:
	\[
	\phi^{(i-1)N}(x) = \begin{cases}
		+k, & x \in \Omega_i, \; \phi(P_i)<0, \\
		-k, & x \in \Omega_i, \; \phi(P_i)>0, \\
		\phi^{(i-1)N}(x), & \text{otherwise}.
	\end{cases}
	\]
	\hspace*{0.5cm} where $k = \frac{\text{Area}(\Omega \setminus \Omega_i)}{\text{Area}(\Omega_i)}$. \\
	\hspace*{0.5cm} \textbf{Modify the velocity field} $F_i$:
	\[
	F_i(x) = \begin{cases}
		a_i, & x \in \Omega_i, \; \phi(P_i)<0, \\
		-a_i, & x \in \Omega_i, \; \phi(P_i)>0, \\
		F_{i-1}(x), & \text{otherwise}.
	\end{cases}
	\]
	\hspace*{0.5cm} \textbf{Repeat $N$ times:} Update the LSF according to Eq. \eqref{iteration}. \\
	\hspace*{0.5cm} \textbf{Output:} The final LSF $\phi^{iN}$ and image with segmentation results. \\
	\textbf{End If}
	
\end{algorithm}

\subsection{Discussions}

Distinguishing itself from convex relaxation techniques that rely on global energy reformulation, SILSM adopts a targeted mechanism to mitigate the susceptibility of standard region-based models (e.g., the CV model) to complex background clutter. Specifically, we introduce an interactive contracting force that effectively constrains the curve evolution within background regions. This mechanism suppresses interference and prevents the contour from becoming trapped in spurious local minima, all while preserving the integrity of the original segmentation core. Consequently, interaction information serves a dual role: primarily, it establishes an initial velocity field to actively curb over-segmentation; secondarily, it provides a robust initialization for the LSF in the vicinity of the target boundary.

Furthermore, we reframe the existence of local minima in variational models not as a defect, but as a representation of intrinsic solution diversity. Navigating the evolution toward a specific, semantically correct minimum requires external priors, which defines the tertiary role of interaction: to guide the model out of an undesirable local minimum toward a user-satisfactory result. To facilitate this transition, we incorporate the MBE regularization term, which ensures that the evolution remains stable and numerically controllable. The synergy between this interactive guidance and stabilized evolution constitutes the complete SILSM framework.

\section{Experiments} \label{sec4}

In this section, we provide a comprehensive evaluation of the proposed SILSM framework. We begin by demonstrating the model's versatility through specific case studies, illustrating its capability in handling distinct interactive scenarios such as sequential multi-object segmentation, fine-grained refinement, and error correction. Following these illustrative examples, we present a quantitative comparative analysis using both medical and real-world datasets, benchmarking our model against established approaches including the Badshah model \cite{badshah2010image}, the Ali model \cite{ali2021image}, and \href{https://github.com/halleewong/ScribblePrompt}{ScribblePrompt} \cite{wong2024scribbleprompt}---an interactive deep learning framework for multi-task medical image segmentation.

For the Badshah and Ali models, the initial LSF was implemented according to the methodology described in their original publications:
\begin{equation}
	\phi_0 = \sqrt{(x-x_0)^2+(y-y_0)^2}-r_0,
\end{equation}
where $(x_0, y_0)$ represents the average of the $x$ and $y$ components of $n$ markers, and the radius $r_0$ is defined as:
\begin{equation}
	r_0 = \min_{\mathbf{y}} \| \mathbf{x}-\mathbf{y} \|,
\end{equation}
where $\mathbf{x} = (x_0, y_0)$ and $\mathbf{y} \in \{(x_i, y_i) \in \Omega, 1 \le i \le n\} \subset \Omega$. Unlike these level-set-based methods, ScribblePrompt employs a bounding-box-based interaction paradigm. In our proposed model, the parameters are set as $\mu=1$ and $\alpha=5$.

The performance of these methods was quantitatively evaluated using several standard metrics: the Dice coefficient, Jaccard index, precision, and recall \cite{chang2009performance,taha2015metrics}. The definitions of these metrics are provided as follows:
$$
\begin{gathered}
	\operatorname{Dice}\left(R_{\text {seg }}, R_{g t}\right)=\frac{2\left|R_{\text {seg }} \cap R_{g t}\right|}{\left|R_{\text {seg }}\right|+\left|R_{g t}\right|} ,\\
	\operatorname{Jaccard}\left(R_{\text {seg }}, R_{g t}\right)=\frac{\left|R_{\text {seg }} \cap R_{g t}\right|}{\left|R_{\text {seg }} \cup R_{g t}\right|} ,\\
	\operatorname{Precision}=\frac{\left|R_{\text {seg }} \cap R_{g t}\right|}{\left|R_{\text {seg }}\right|} ,\\
	\operatorname{Recall}=\frac{\left|R_{\text {seg }} \cap R_{g t}\right|}{\left|R_{\text {gt }}\right|}.
\end{gathered}
$$

\subsection{Illustrative examples of interactive capabilities}
To highlight the flexibility and responsiveness of the proposed method, we present three distinct operational cases. These examples demonstrate how the interaction term $F$ and the speed parameter $a_i$ effectively facilitate progressive refinement and error correction. Specifically, we illustrate the algorithm's capability in segmenting multiple objects sequentially, achieving finer segmentation of internal structures, and correcting inaccurate results. In the following experiments, the default parameters are set to $\mu=1$, $\alpha=5$, time step $\Delta t=0.1$, and iterations $N=200$, with the speed parameter $a_1=500$.

\subsubsection*{Case 1: Segmenting multiple objects sequentially}
\begin{figure}[!ht]
    \centering
    \includegraphics[width=2in]{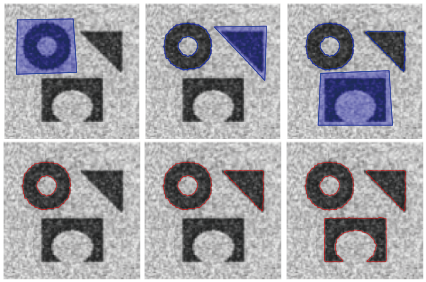}
    \caption{Top: Sequence of interactively added regions of interest. Bottom: Progressive segmentation results after each interaction.}
    \label{fig6}
\end{figure}

The process is demonstrated using a synthetic image containing noise. Fig. \ref{fig6} displays the selected region of interest alongside the segmentation result for each interaction step. By sequentially defining the region of interest, the model effectively isolates multiple objects one by one without interference. In this experiment, the speed parameter is set to $a_i=0$ for $i\ge2$.

\subsubsection*{Case 2: Achieving finer segmentation of objects}
\begin{figure}[!ht]
    \centering
    \includegraphics[width=3in]{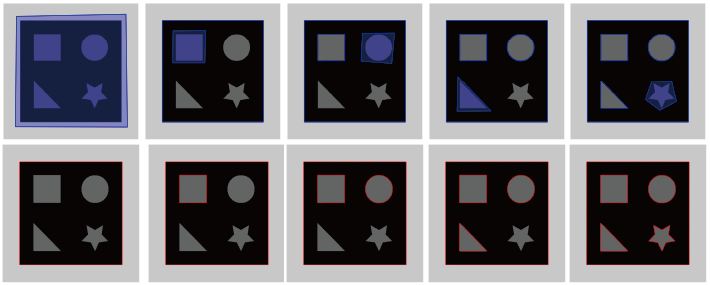}
    \caption{Top: Sequence of interactively added regions of interest. Bottom: Progressive segmentation results after each interaction.}
    \label{fig7}
\end{figure}

We employ a synthetic image with nested structures to illustrate this task. As shown in Fig. \ref{fig7}, the initial global segmentation fails to resolve the separate shapes inside the black rectangle. Subsequently, through interactive refinement (adding exclusion/inclusion zones), these internal shapes are successfully segmented individually. In this experiment, the parameter is set to $a_i=20$ for $i\ge2$ to ensure strong background suppression in the refined regions.

\subsubsection*{Case 3: Correcting inaccurate segmentation results}
\begin{figure}[!ht]
    \centering
    \includegraphics[width=3.5in]{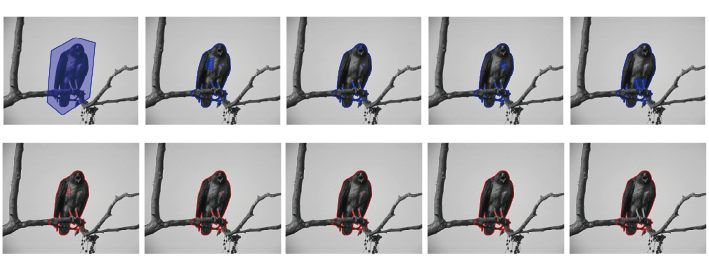}
    \caption{The process of modifying the segmentation results involves outlining the desired area for adjustment in each step.}
    \label{fig8}
\end{figure}

We employ a real-world image to demonstrate our model's capability to rectify segmentation errors. Owing to the fixed coefficients preceding the data fidelity term, the initial segmentation is often suboptimal (e.g., leaking into the background or missing parts of the object). Fig. \ref{fig8} illustrates the sequential process of interactively correcting an inadequate result to achieve the desired segmentation. By specifying the error region, the model updates the contour locally. In this experiment, the parameter is set to $a_i=0$ for $i\ge2$.

\subsection{Medical image segmentation}

\begin{figure}[!ht]
	\centering
	
	\subfloat[]{\includegraphics[width=0.24\textwidth]{9.1.png}}
	\hspace{0.01\textwidth} 
	\subfloat[]{\includegraphics[width=0.24\textwidth]{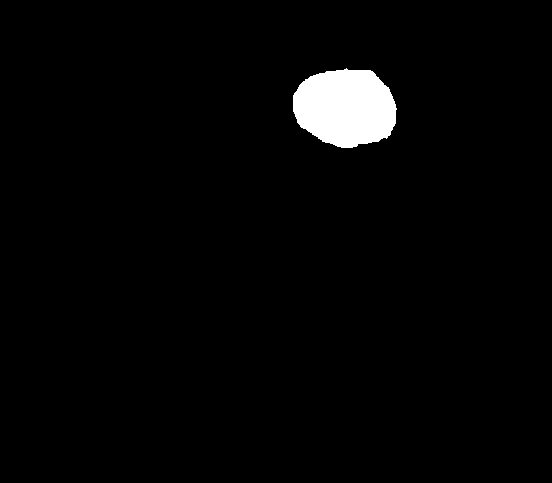}}
	\hspace{0.01\textwidth}
	\subfloat[]{\includegraphics[width=0.24\textwidth]{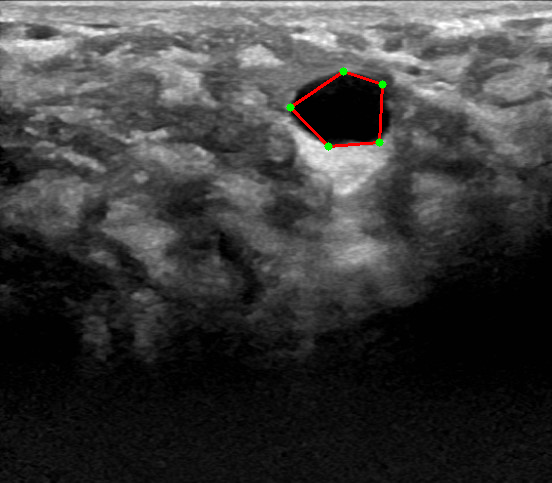}}
	
	\vspace{0.3em} 
	
	\subfloat[]{\includegraphics[width=0.24\textwidth]{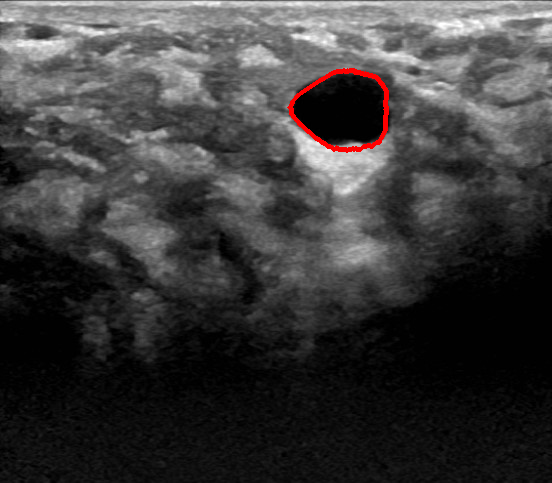}}
	\hspace{0.01\textwidth}
	\subfloat[]{\includegraphics[width=0.24\textwidth]{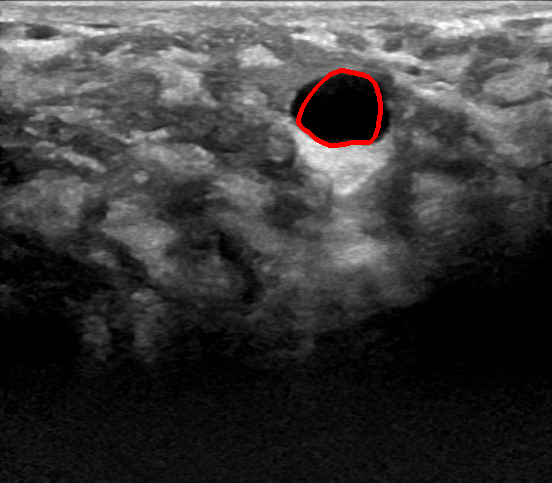}}
	\hspace{0.01\textwidth}
	\subfloat[]{\includegraphics[width=0.24\textwidth]{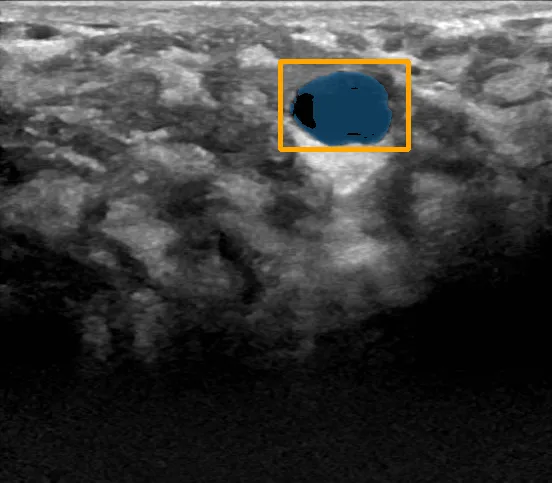}}
	
	\vspace{0.3em}
	
	\subfloat[]{\includegraphics[width=0.24\textwidth]{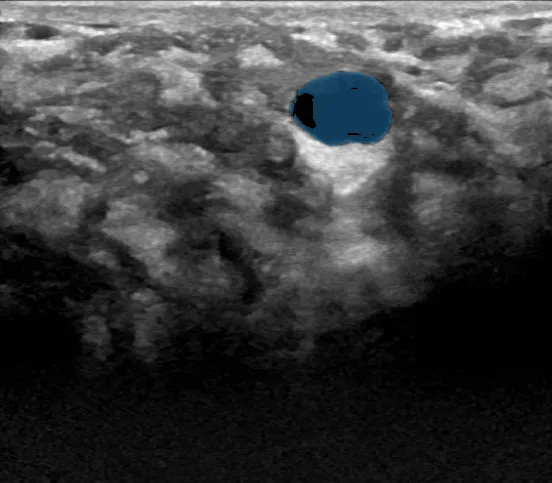}}
	\hspace{0.01\textwidth}
	\subfloat[]{\includegraphics[width=0.24\textwidth]{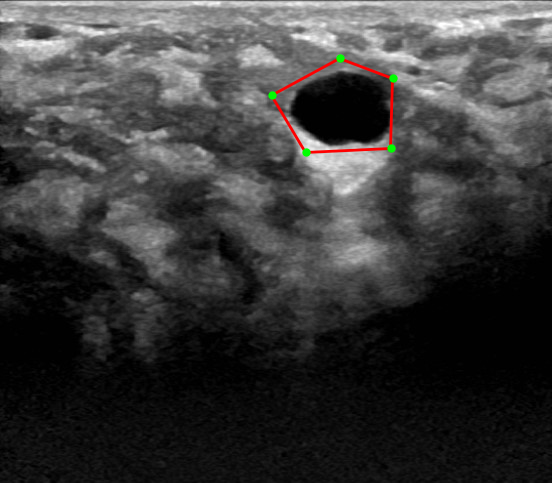}}
	\hspace{0.01\textwidth}
	\subfloat[]{\includegraphics[width=0.24\textwidth]{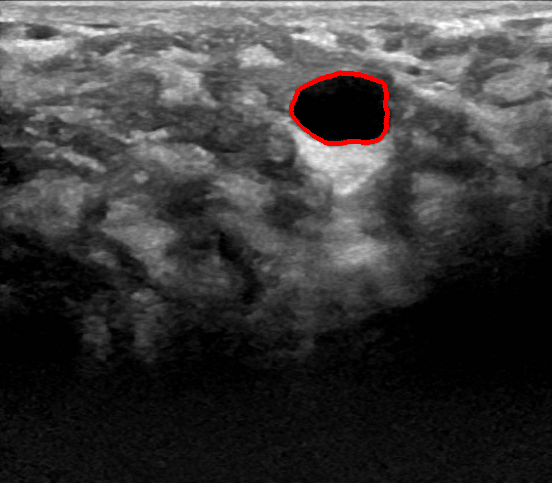}}
	
	\caption{Qualitative comparison of interactive segmentation results on a breast ultrasound case. (a) Original ultrasound image. (b) Manual ground truth annotation. (c) Initial interactive scribbles used for the Badshah and Ali models. (d) Segmentation result of the Badshah model. (e) Segmentation result of the Ali model. (f) Bounding-box interaction for ScribblePrompt. (g) Segmentation result of ScribblePrompt. (h) Initial interactive scribbles used for our proposed model. (i) Final segmentation result of our proposed model.}
	\label{fig9}
\end{figure}

To assess the segmentation performance of the proposed model on medical imaging tasks, we conducted a visual evaluation using representative breast ultrasound images and lung CT scans. Fig.~\ref{fig9} illustrates the segmentation results for a breast ultrasound image, displaying the selected interaction points alongside the corresponding outputs for all evaluated models. Table~\ref{Tab1} summarizes the quantitative metrics, with the best performance in each category highlighted in bold.

\begin{table}[!ht] 
	\centering 
	\caption{Quantitative comparisons of our method and other methods.}
	\label{Tab1} 
	\begin{tabular*}{\textwidth}{@{\extracolsep{\fill}}lcccc} 
		\toprule
		Method & Dice & Jaccard & Precision & Recall \\
		\midrule
		Badshah et al. & 0.9111 & 0.8367 & 0.9650 & \textbf{0.8629} \\
		Ali et al. & 0.8348 & 0.7165 & 0.9890 & 0.7223 \\
		ScribblePrompt & 0.8596 & 0.7538 & 0.9803 & 0.7654 \\
		\textbf{Our method} & \textbf{0.9186} & \textbf{0.8495} & \textbf{0.9887} & 0.8578 \\
		\bottomrule
	\end{tabular*}
\end{table}

As illustrated, the contour produced by the Badshah model fails to accurately converge to the target boundary and exhibits a lack of smoothness characterized by noticeable structural irregularities. Constrained by its area term, the Ali model requires a significant number of interaction points to approximate a near-circular target with a polygonal representation; an insufficient number of points results in a contour that deviates substantially from the ground truth. While ScribblePrompt facilitates iterative refinement through subsequent interactions, its initial segmentation result in this instance remains suboptimal. In contrast, our model achieves a satisfactory segmentation result with only the initial interaction. Subsequently, a lung CT case is presented to demonstrate the capability of our model to progressively refine segmentation outputs through additional user interactions when initial results are insufficient.

\begin{figure}[!ht]
    \centering
    
    \subfloat[]{\includegraphics[width=0.24\textwidth]{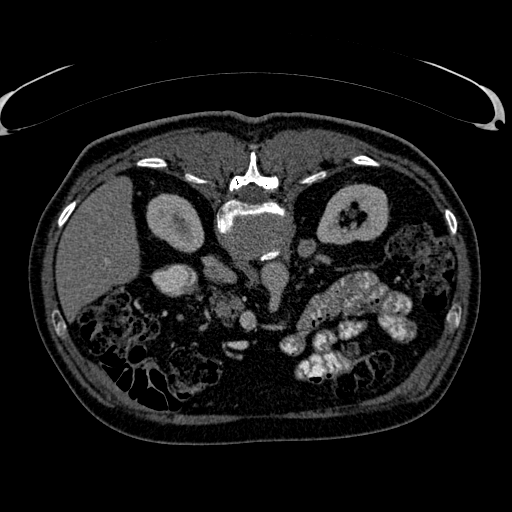}}
    \hspace{0.01\textwidth} 
    \subfloat[]{\includegraphics[width=0.24\textwidth]{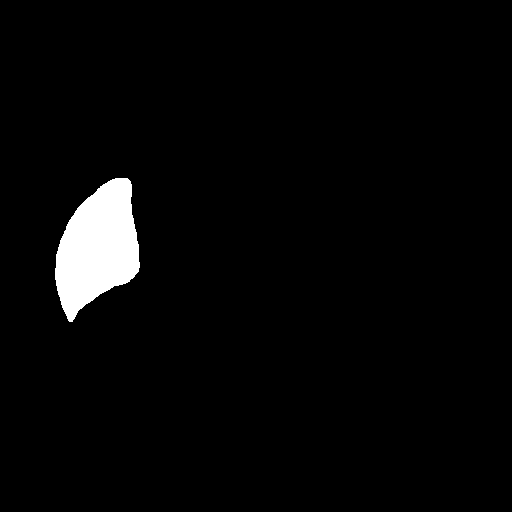}}
    \hspace{0.01\textwidth}
    \subfloat[]{\includegraphics[width=0.24\textwidth]{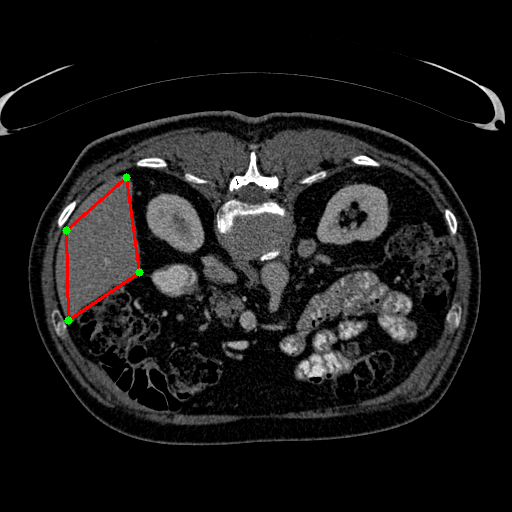}}
    
    \vspace{0.3em} 
   
    \subfloat[]{\includegraphics[width=0.24\textwidth]{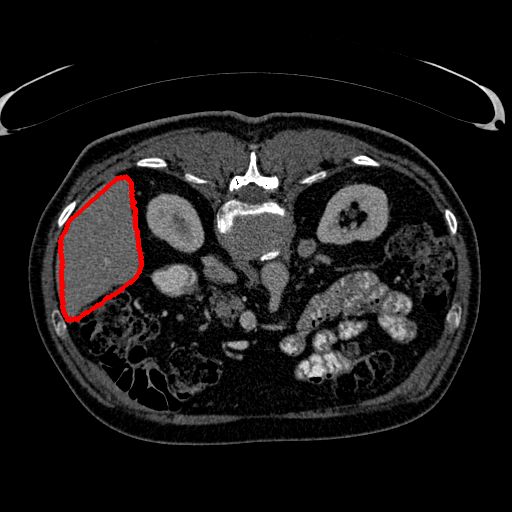}}
    \hspace{0.01\textwidth}
    \subfloat[]{\includegraphics[width=0.24\textwidth]{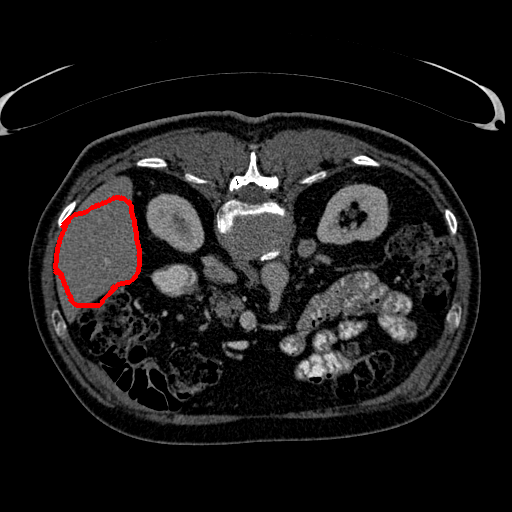}}
    \hspace{0.01\textwidth}
    \subfloat[]{\includegraphics[width=0.24\textwidth]{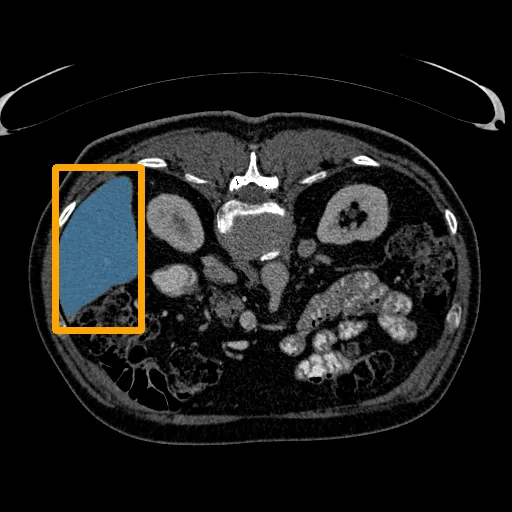}}
    
    \vspace{0.3em}
    
    \subfloat[]{\includegraphics[width=0.24\textwidth]{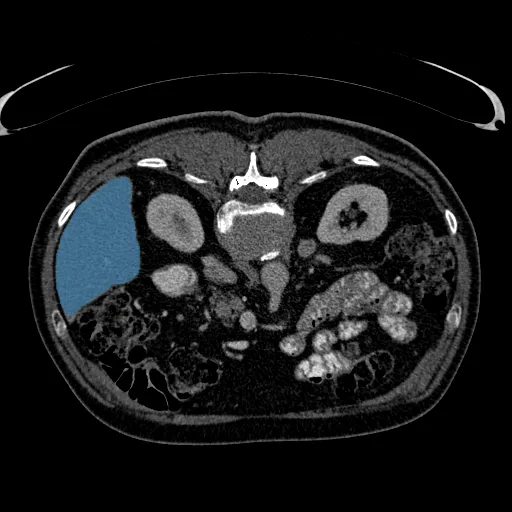}}
    \hspace{0.01\textwidth}
    \subfloat[]{\includegraphics[width=0.24\textwidth]{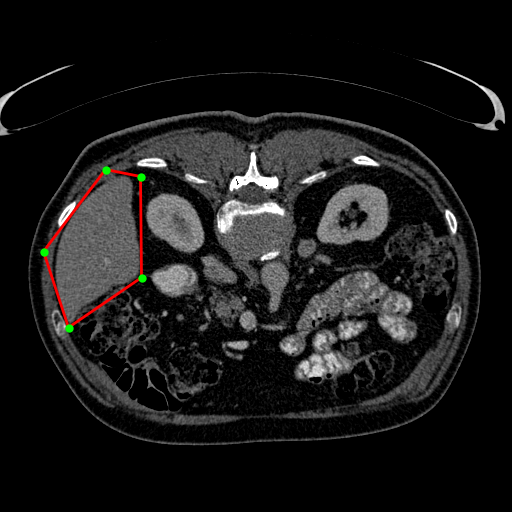}}
    \hspace{0.01\textwidth}
    \subfloat[]{\includegraphics[width=0.24\textwidth]{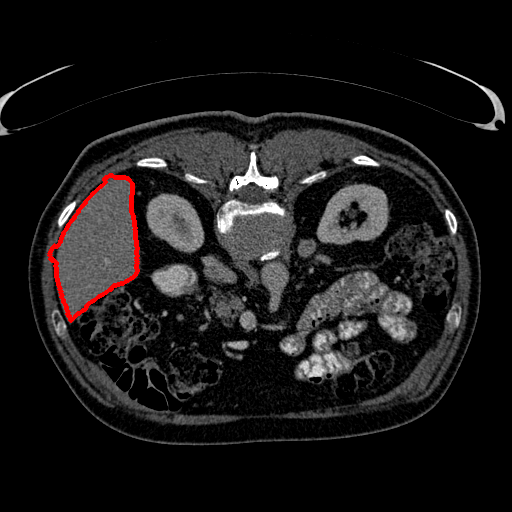}}
    
    \caption{Qualitative comparison of interactive segmentation results on a lung CT image. (a) Original CT image. (b) Manual ground truth annotation. (c) Initial interactive scribbles used for the Badshah and Ali models. (d) Segmentation result of the Badshah model. (e) Segmentation result of the Ali model. (f) Bounding-box interaction for ScribblePrompt. (g) Segmentation result of ScribblePrompt. (h) Initial interactive scribbles used for our proposed model. (i) Final segmentation result of our proposed model.}
    \label{fig10}
\end{figure}

Fig.~\ref{fig10} illustrates the segmentation results for the lung CT image, presenting the selected interaction points alongside the corresponding outputs for all competitive models. As observed, the initial contour contains two prominent convex protrusions. These geometrically inconsistent regions were subsequently rectified through two incremental interaction steps. 

\begin{figure}[!ht]
	\centering
	\includegraphics[width=0.8\textwidth]{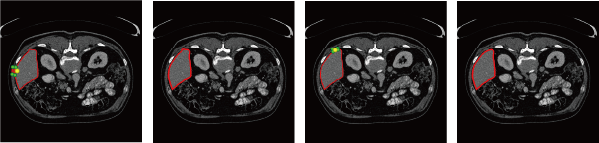} 
	\caption{Two additional interaction steps along with the segmentation results after each step.}
	\label{fig11}
\end{figure}

\begin{table}[htbp] 
	\centering 
	\caption{Quantitative comparisons of our method and other methods.}
	\label{Tab2} 
	\begin{tabular*}{\textwidth}{@{\extracolsep{\fill}}lcccc} 
		\toprule
		Method & Dice & Jaccard & Precision & Recall \\
		\midrule
		Badshah et al. & 0.9239 & 0.8585 & 0.9271 & 0.9207 \\
		Ali et al. & 0.8808 & 0.7870 & 0.9651 & 0.8101 \\
		\textbf{ScribblePrompt} & \textbf{0.9805} & \textbf{0.9618} & \textbf{0.9900} & \textbf{0.9713} \\
		Our for 1 interaction & 0.9656 & 0.9335 & 0.9647 & 0.9620 \\
		Our for 2 interaction & 0.9711 & 0.9439 & 0.9772 & 0.9651 \\
		\textbf{Our for 3 interaction} & \textbf{0.9731} & \textbf{0.9476} & \textbf{0.9845} & \textbf{0.9665} \\
		\bottomrule
	\end{tabular*}
\end{table}

The refinement process is detailed in Fig.~\ref{fig11}, where yellow points denote the interaction points $P_i$ used for region classification in Algorithm~\ref{alg1}, and green points represent the markers delineating the area of interest $\Omega_i$. As the interaction progresses, initial inaccuracies are progressively eliminated, ultimately yielding an optimal segmentation. Table~\ref{Tab2} lists the quantitative metrics for all models, with the best and second-best results highlighted in bold.

For each refinement step, the parameter was fixed at $a_i=0$. It is observed that all evaluation metrics improve monotonically as the interaction proceeds, eventually achieving a performance level comparable to that of state-of-the-art deep neural networks.

\subsection{Real-world image segmentation}

Beyond medical imaging, our model demonstrates robust generalization capabilities in the challenging domain of real-world image segmentation. Unlike conventional interactive methods---which often struggle with complex geometries and demand a high density of precisely positioned markers---our approach significantly alleviates the user burden; a single bounding box is sufficient to initialize the segmentation. Furthermore, should refinements be necessary, the model supports iterative corrections without requiring a complete restart, thereby facilitating a more intuitive and efficient workflow. The following examples further illustrate these practical advantages.

\begin{figure}[!ht]
	\centering
	\subfloat[]{\includegraphics[width=0.24\textwidth]{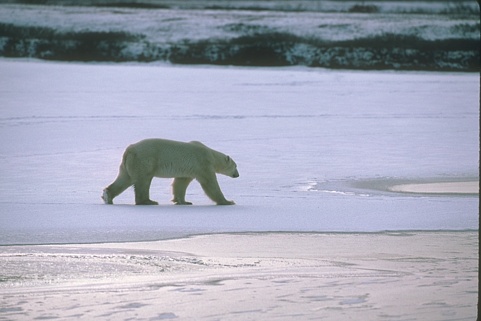}}
	\hfill
	\subfloat[]{\includegraphics[width=0.24\textwidth]{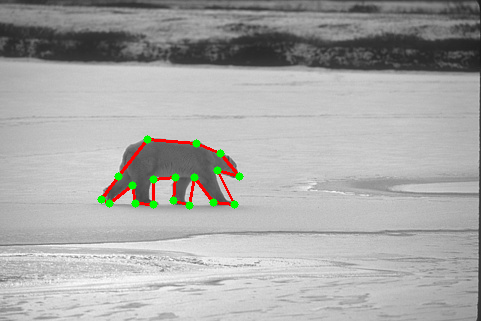}}
	\hfill
	\subfloat[]{\includegraphics[width=0.24\textwidth]{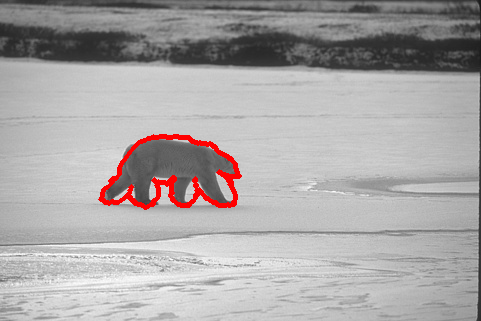}}
	\hfill
	\subfloat[]{\includegraphics[width=0.24\textwidth]{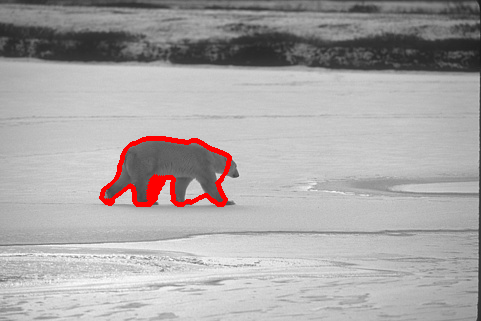}}
	
	\subfloat[]{\includegraphics[width=0.24\textwidth]{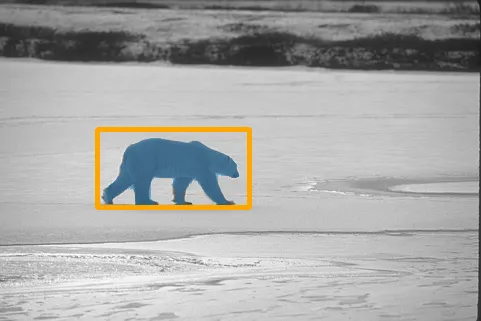}}
	\hfill
	\subfloat[]{\includegraphics[width=0.24\textwidth]{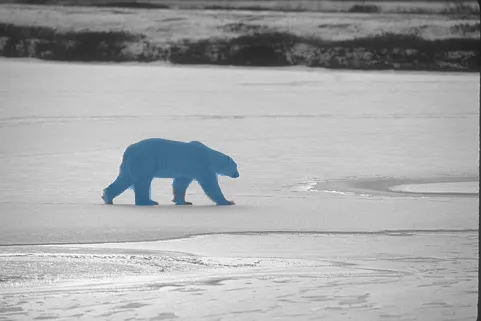}}
	\hfill
	\subfloat[]{\includegraphics[width=0.24\textwidth]{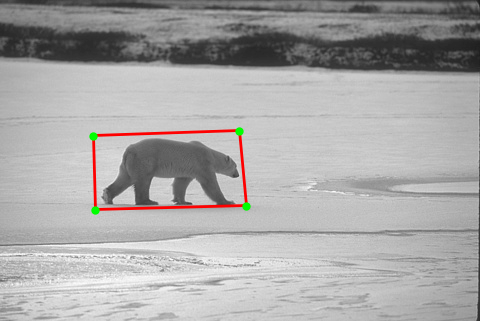}}
	\hfill
	\subfloat[]{\includegraphics[width=0.24\textwidth]{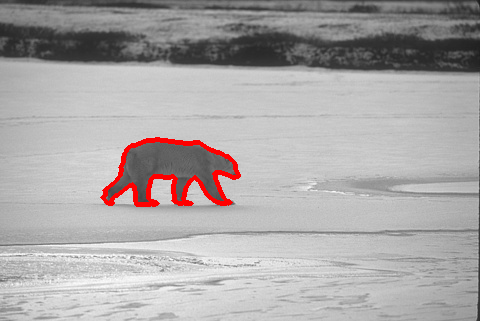}}
	
	\caption{(a) Original image. (b) Initial interactive scribbles used for the Badshah and Ali models. (c) Segmentation result of the Badshah model. (d) Segmentation result of the Ali model. (e) Bounding-box interaction for ScribblePrompt. (f) Segmentation result of ScribblePrompt. (g) Initial interactive scribbles used for our proposed model. (h)Final segmentation result of our proposed model.}
	\label{fig12}
\end{figure}

As illustrated in Fig. \ref{fig12}, even with an adequate number of interaction points, both the Badshah and Ali models struggle to accurately delineate the target object boundaries. ScribblePrompt, a deep neural network specifically tailored for medical imaging, demonstrates a certain degree of cross-domain robustness when applied to real-world images. However, it exhibits significant limitations in capturing fine-grained structural details, as evidenced by the imprecise segmentation of the target's leg region in Fig. \ref{fig12}. In contrast, our model yields a superior initial segmentation, outperforming all aforementioned approaches in terms of both boundary adherence and topological accuracy.

\begin{figure}[!ht]
	\centering
	\subfloat[]{\includegraphics[width=0.2\textwidth]{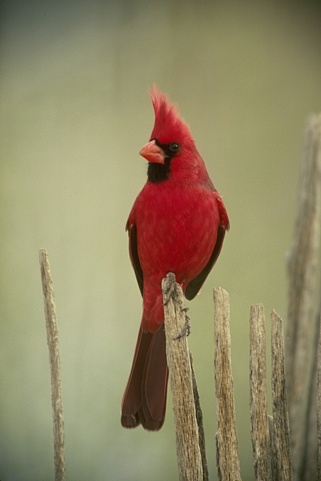}}
	\hfill
	\subfloat[]{\includegraphics[width=0.2\textwidth]{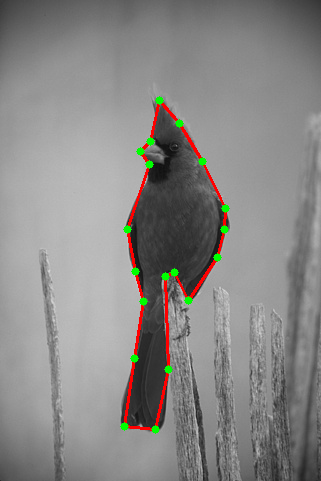}}
	\hfill
	\subfloat[]{\includegraphics[width=0.2\textwidth]{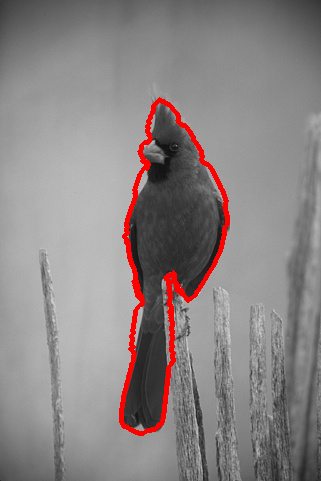}}
	\hfill
	\subfloat[]{\includegraphics[width=0.2\textwidth]{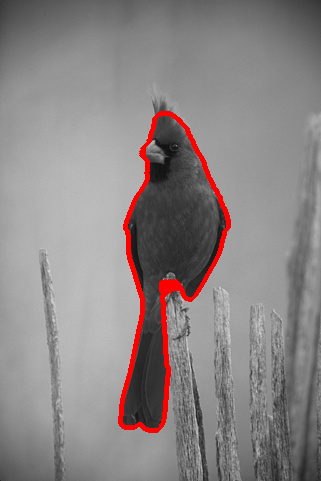}} 
	
	\subfloat[]{\includegraphics[width=0.2\textwidth]{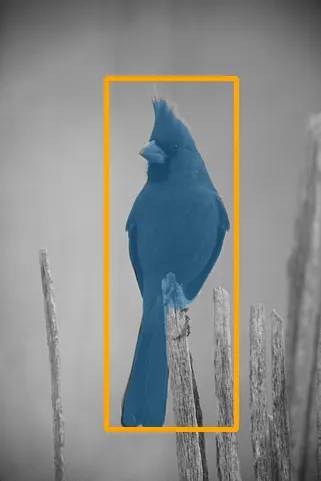}}
	\hfill
	\subfloat[]{\includegraphics[width=0.2\textwidth]{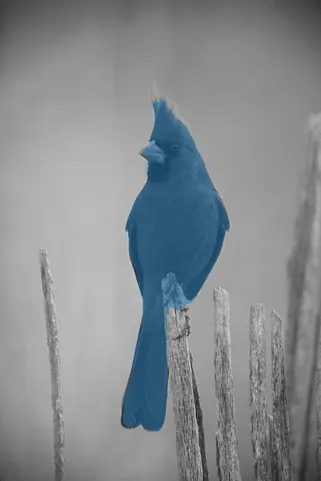}}
	\hfill
	\subfloat[]{\includegraphics[width=0.2\textwidth]{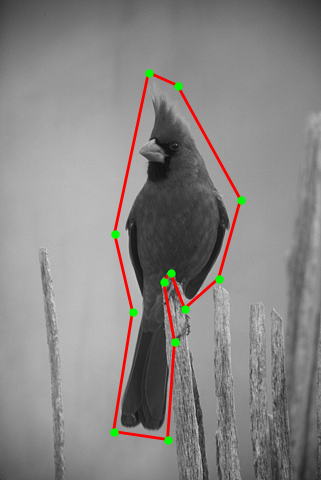}}
	\hfill
	\subfloat[]{\includegraphics[width=0.2\textwidth]{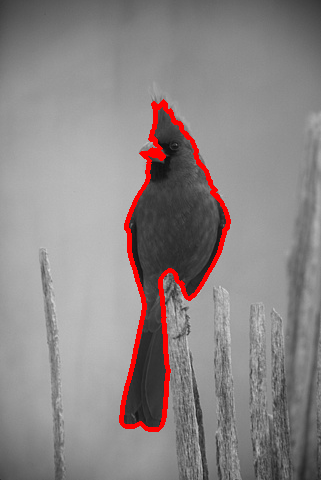}}
	\caption{(a) Original image. (b) Initial interactive scribbles used for the Badshah and Ali models. (c) Segmentation result of the Badshah model. (d) Segmentation result of the Ali model. (e) Bounding-box interaction for ScribblePrompt. (f) Segmentation result of ScribblePrompt. (g) Initial interactive scribbles used for our proposed model. (h)Final segmentation result of our proposed model.}
	\label{fig13}
\end{figure}

\begin{figure}[!ht]
	\centering
	\includegraphics[width=\linewidth]{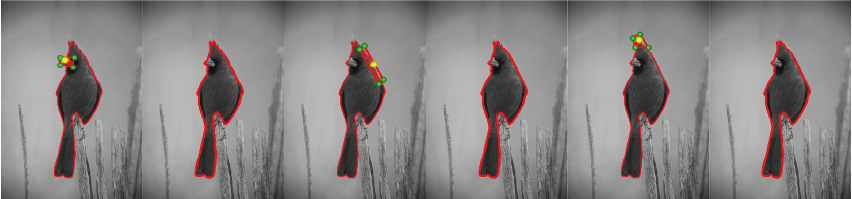}
	\caption{Three rounds of interaction and the corresponding segmentation results.}
	\label{fig14}
\end{figure}

For complex scenarios where a single interaction may produce suboptimal results, our model facilitates an iterative refinement mechanism, similar to the process shown in Fig. \ref{fig11}. Fig. \ref{fig13} presents the interactive inputs alongside the corresponding segmentation outcomes for all competitive models. Furthermore, Fig.~\ref{fig14} visualizes the refinement trajectory of our model, demonstrating how segmentation quality is monotonically enhanced through successive, low-effort interactions.

\subsection{Stability of multiple segmentation}

\begin{figure}[!t]
	\centering
	
	\subfloat[]{\includegraphics[width=0.24\textwidth]{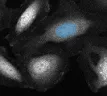}}
	\hspace{0.01\textwidth} 
	\subfloat[]{\includegraphics[width=0.24\textwidth]{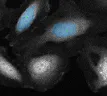}}
	\hspace{0.01\textwidth}
	\subfloat[]{\includegraphics[width=0.24\textwidth]{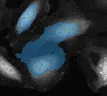}}
	
	\vspace{0.3em} 
	
	\subfloat[]{\includegraphics[width=0.24\textwidth]{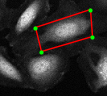}}
	\hspace{0.01\textwidth}
	\subfloat[]{\includegraphics[width=0.24\textwidth]{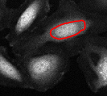}}
	\hspace{0.01\textwidth}
	\subfloat[]{\includegraphics[width=0.24\textwidth]{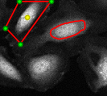}}
	
	\vspace{0.3em}
	
	\subfloat[]{\includegraphics[width=0.24\textwidth]{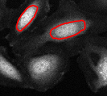}}
	\hspace{0.01\textwidth}
	\subfloat[]{\includegraphics[width=0.24\textwidth]{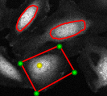}}
	\hspace{0.01\textwidth}
	\subfloat[]{\includegraphics[width=0.24\textwidth]{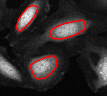}}
	
	\caption{(a)-(c) The segmentation results of ScribblePrompt after each of the three interactions. (d)-(i) The interactive approach of our proposed model over three iterations and the segmentation results obtained after each interaction.}
	\label{fig15}
\end{figure}

We employ a multi-cell segmentation task to evaluate the stability of the proposed method relative to the deep neural network, ScribblePrompt, across multiple interaction cycles. ScribblePrompt utilizes an interaction paradigm based on inclusion points. As illustrated in Fig.~\ref{fig15}, the experimental results reveal that ScribblePrompt's performance becomes unstable following the third interaction, exhibiting significant structural deviations from the expected regions. In contrast, our model consistently maintains robust and stable performance throughout the entire interactive sequence, demonstrating superior reliability in iterative tasks.

\section{Conclusion} \label{sec5}

In this work, we move beyond the conventional one-shot interaction paradigm and introduce a novel framework termed Sustainable Interactive Segmentation, which enables stable and continuous refinement through ongoing user interaction.
To realize this concept, we propose the SILSM, a flexible and general variational framework that can be seamlessly integrated into a broad class of segmentation models based on level sets. Extensive experiments on diverse medical imaging modalities and real-world scenarios demonstrate the effectiveness and robustness of the proposed approach. In particular, SILSM significantly reduces the number of required interaction cues and exhibits strong insensitivity to inaccurate interactive input.
Moreover, throughout the iterative interaction process, the proposed method maintains a stable evolution and consistently improves the segmentation result, progressively converging toward an accurate and reliable final delineation. These results highlight the potential of sustainable interaction as a promising paradigm for interactive segmentation tasks.
Future work will explore the extension of the proposed framework to learning-based and high-dimensional segmentation settings.

\backmatter

\section*{Acknowledgements} This work is partially supported by the National Natural Science Foundation of China(12171123, 12271130, 12371419, 12501586, U21B2075), National Key R\&D Program of China(2023YFC2205900, 2023YFC2205903).


\bibliography{sn-bibliography}

@article{chan2001active,
	title={Active contours without edges},
	author={Chan, Tony F and Vese, Luminita A},
	journal={IEEE Transactions on image processing},
	volume={10},
	number={2},
	pages={266--277},
	year={2001},
	publisher={IEEE}
}

@article{caselles1997geodesic,
	title={Geodesic active contours},
	author={Caselles, Vicent and Kimmel, Ron and Sapiro, Guillermo},
	journal={International journal of computer vision},
	volume={22},
	number={1},
	pages={61--79},
	year={1997},
	publisher={Springer}
}

@article{mumford1989optimal,
	title={Optimal approximations by piecewise smooth functions and associated variational problems},
	author={Mumford, David and Shah, Jayant},
	journal={Communications on Pure and Applied Mathematics},  
	volume={42},
	number={5},
	pages={577--685},
	year={1989},
	publisher={Wiley Online Library}
}

@article{grady2006random,
	title={Random walks for image segmentation},
	author={Grady, Leo},
	journal={IEEE transactions on pattern analysis and machine intelligence},
	volume={28},
	number={11},
	pages={1768--1783},
	year={2006},
	publisher={IEEE}
}

@inproceedings{bai2007geodesic,
	title={A geodesic framework for fast interactive image and video segmentation and matting},
	author={Bai, Xue and Sapiro, Guillermo},
	booktitle={2007 IEEE 11th International Conference on Computer Vision},
	pages={1--8},
	year={2007},
	organization={IEEE}
}

@article{rother2004grabcut,
	title={" GrabCut" interactive foreground extraction using iterated graph cuts},
	author={Rother, Carsten and Kolmogorov, Vladimir and Blake, Andrew},
	journal={ACM transactions on graphics (TOG)},
	volume={23},
	number={3},
	pages={309--314},
	year={2004},
	publisher={ACM New York, NY, USA}
}

@article{gout2005segmentation,
	title={Segmentation under geometrical conditions using geodesic active contours and interpolation using level set methods},
	author={Gout, Christian and Le Guyader, Carole and Vese, Luminita},
	journal={Numerical algorithms},
	volume={39},
	number={1},
	pages={155--173},
	year={2005},
	publisher={Springer}
}

@article{badshah2010image,
	title={Image selective segmentation under geometrical constraints using an active contour approach},
	author={Badshah, Noor and Chen, Ke},
	journal={Communications in Computational Physics},
	volume={7},
	number={4},
	pages={759},
	year={2010}
}

@article{rada2012new,
	title={A new variational model with dual level set functions for selective segmentation},
	author={Rada, Lavdie and Chen, Ke},
	journal={Communications in Computational Physics},
	volume={12},
	number={1},
	pages={261--283},
	year={2012},
	publisher={Cambridge University Press}
}

@article{rada2013improved,
	title={Improved selective segmentation model using one level-set},
	author={Rada, Lavdie and Chen, Ke},
	journal={Journal of Algorithms \& Computational Technology},
	volume={7},
	number={4},
	pages={509--540},
	year={2013},
	publisher={SAGE Publications Sage UK: London, England}
}

@article{ali2021image,
	title={Image-selective segmentation model for multi-regions within the object of interest with application to medical disease},
	author={Ali, Haider and Faisal, Shah and Chen, Ke and Rada, Lavdie},
	journal={The Visual Computer},
	volume={37},
	number={5},
	pages={939--955},
	year={2021},
	publisher={Springer}
}

@inproceedings{rada2018selective,
	title={A selective segmentation model for inhomogeneous images},
	author={Rada, Lavdie and Ali, Haider and Khan, Haroon Niaz Ali},
	booktitle={International Symposium on Intelligent Computing Systems},
	pages={123--136},
	year={2018},
	organization={Springer}
}

@article{song2024re,
	title={Re-initialization-free level set method via molecular beam epitaxy equation regularization for image segmentation},
	author={Song, Fanghui and Sun, Jiebao and Shi, Shengzhu and Guo, Zhichang and Zhang, Dazhi},
	journal={Journal of Mathematical Imaging and Vision},
	volume={66},
	number={5},
	pages={926--950},
	year={2024},
	publisher={Springer}
}

@article{le2008geodesic,
	title={Geodesic active contour under geometrical conditions: Theory and 3D applications},
	author={Le Guyader, Carole and Gout, Christian},
	journal={Numerical algorithms},
	volume={48},
	number={1},
	pages={105--133},
	year={2008},
	publisher={Springer}
}

@article{zhao2022new,
	title={A new variational method for selective segmentation of medical images},
	author={Zhao, Wenxiu and Wang, Weiwei and Feng, Xiangchu and Han, Yu},
	journal={Signal Processing},
	volume={190},
	pages={108292},
	year={2022},
	publisher={Elsevier}
}

@article{zhang2017global,
	title={Global sparse gradient guided variational Retinex model for image enhancement},
	author={Zhang, Rui and Feng, Xiangchu and Yang, Lixia and Chang, Lihong and Xu, Chen},
	journal={Signal Processing: Image Communication},
	volume={58},
	pages={270--281},
	year={2017},
	publisher={Elsevier}
}

@inproceedings{jain2023oneformer,
	title={Oneformer: One transformer to rule universal image segmentation},
	author={Jain, Jitesh and Li, Jiachen and Chiu, Mang Tik and Hassani, Ali and Orlov, Nikita and Shi, Humphrey},
	booktitle={Proceedings of the IEEE/CVF conference on computer vision and pattern recognition},
	pages={2989--2998},
	year={2023}
}

@article{allabadi2025learning,
  title={Learning to detect novel species with SAM in the wild},
  author={Allabadi, Garvita and Lucic, Ana and Wang, Yu-Xiong and Adve, Vikram},
  journal={International Journal of Computer Vision},
  volume={133},
  number={5},
  pages={2247--2258},
  year={2025},
  publisher={Springer}
}

@article{osher1988fronts,
	title={Fronts propagating with curvature-dependent speed: Algorithms based on Hamilton-Jacobi formulations},
	author={Osher, Stanley and Sethian, James A},
	journal={Journal of computational physics},
	volume={79},
	number={1},
	pages={12--49},
	year={1988},
	publisher={Elsevier}
}

@inproceedings{wong2024scribbleprompt,
	title={Scribbleprompt: fast and flexible interactive segmentation for any biomedical image},
	author={Wong, Hallee E and Rakic, Marianne and Guttag, John and Dalca, Adrian V},
	booktitle={European Conference on Computer Vision},
	pages={207--229},
	year={2024},
	organization={Springer}
}

@article{chang2009performance,
	title={Performance measure characterization for evaluating neuroimage segmentation algorithms},
	author={Chang, Herng-Hua and Zhuang, Audrey H and Valentino, Daniel J and Chu, Woei-Chyn},
	journal={Neuroimage},
	volume={47},
	number={1},
	pages={122--135},
	year={2009},
	publisher={Elsevier}
}

@article{taha2015metrics,
	title={Metrics for evaluating 3D medical image segmentation: analysis, selection, and tool},
	author={Taha, Abdel Aziz and Hanbury, Allan},
	journal={BMC medical imaging},
	volume={15},
	number={1},
	pages={29},
	year={2015},
	publisher={Springer}
}

@article{spencer2015convex,
	title={A convex and selective variational model for image segmentation},
	author={Spencer, Jack and Chen, Ke},
	journal={Communications in Mathematical Sciences},
	volume={13},
	number={6},
	pages={1453--1472},
	year={2015}
}

@inproceedings{kirillov2023segment,
	title={Segment anything},
	author={Kirillov, Alexander and Mintun, Eric and Ravi, Nikhila and Mao, Hanzi and Rolland, Chloe and Gustafson, Laura and Xiao, Tete and Whitehead, Spencer and Berg, Alexander C and Lo, Wan-Yen and others},
	booktitle={Proceedings of the IEEE/CVF international conference on computer vision},
	pages={4015--4026},
	year={2023}
}

@article{luo2021mideepseg,
	title={MIDeepSeg: Minimally interactive segmentation of unseen objects from medical images using deep learning},
	author={Luo, Xiangde and Wang, Guotai and Song, Tao and Zhang, Jingyang and Aertsen, Michael and Deprest, Jan and Ourselin, Sebastien and Vercauteren, Tom and Zhang, Shaoting},
	journal={Medical image analysis},
	volume={72},
	pages={102102},
	year={2021},
	publisher={Elsevier}
}

@inproceedings{liu2023simpleclick,
	title={Simpleclick: Interactive image segmentation with simple vision transformers},
	author={Liu, Qin and Xu, Zhenlin and Bertasius, Gedas and Niethammer, Marc},
	booktitle={Proceedings of the IEEE/CVF International Conference on Computer Vision},
	pages={22290--22300},
	year={2023}
}

@article{yu2023techniques,
	title={Techniques and challenges of image segmentation: A review},
	author={Yu, Ying and Wang, Chunping and Fu, Qiang and Kou, Renke and Huang, Fuyu and Yang, Boxiong and Yang, Tingting and Gao, Mingliang},
	journal={Electronics},
	volume={12},
	number={5},
	pages={1199},
	year={2023},
	publisher={MDPI}
}

@article{zhang2025small,
	title={Small object few-shot segmentation for vision-based industrial inspection},
	author={Zhang, Zilong and Niu, Chang and Zhao, Zhibin and Zhang, Xingwu and Chen, Xuefeng},
	journal={IEEE Transactions on Industrial Informatics},
	year={2025},
	publisher={IEEE}
}

@article{liu2023survey,
	title={A survey of real-time surface defect inspection methods based on deep learning},
	author={Liu, Yi and Zhang, Changsheng and Dong, Xingjun},
	journal={Artificial Intelligence Review},
	volume={56},
	number={10},
	pages={12131--12170},
	year={2023},
	publisher={Springer}
}

@article{ramesh2021review,
	title={A review of medical image segmentation algorithms.},
	author={Ramesh, KKD and Kumar, G Kiran and Swapna, K and Datta, Debabrata and Rajest, S Suman},
	journal={EAI Endorsed Transactions on Pervasive Health \& Technology},
	volume={7},
	number={27},
	year={2021}
}

@article{feng2020deep,
	title={Deep multi-modal object detection and semantic segmentation for autonomous driving: Datasets, methods, and challenges},
	author={Feng, Di and Haase-Sch{\"u}tz, Christian and Rosenbaum, Lars and Hertlein, Heinz and Glaeser, Claudius and Timm, Fabian and Wiesbeck, Werner and Dietmayer, Klaus},
	journal={IEEE Transactions on Intelligent Transportation Systems},
	volume={22},
	number={3},
	pages={1341--1360},
	year={2020},
	publisher={IEEE}
}

@article{sun2025efficient,
  title={On Efficient Variants of Segment Anything Model: A Survey: X. Sun et al.},
  author={Sun, Xiaorui and Liu, Jun and Shen, Hengtao and Zhu, Xiaofeng and Hu, Ping},
  journal={International Journal of Computer Vision},
  volume={133},
  number={10},
  pages={7406--7436},
  year={2025},
  publisher={Springer}
}

@article{bogensperger2025flowsdf,
  title={FlowSDF: Flow matching for medical image segmentation using distance transforms},
  author={Bogensperger, Lea and Narnhofer, Dominik and Falk, Alexander and Schindler, Konrad and Pock, Thomas},
  journal={International Journal of Computer Vision},
  pages={1--13},
  year={2025},
  publisher={Springer}
}

@book{evans2022partial,
  title={Partial differential equations},
  author={Evans, Lawrence C},
  volume={19},
  year={2022},
  publisher={American Mathematical Society},
  address={Providence, RI},
}

@article{vobecky2025unsupervised,
  title={Unsupervised semantic segmentation of urban scenes via cross-modal distillation},
  author={Vobecky, Antonin and Hurych, David and Simeoni, Oriane and Gidaris, Spyros and Bursuc, Andrei and Perez, Patrick and Sivic, Josef},
  journal={International Journal of Computer Vision},
  volume={133},
  number={6},
  pages={3519--3541},
  year={2025},
  publisher={Springer}
}

@book{Attouch2014,
	author    = {Attouch, H. and Buttazzo, G. and Michaille, G.},
	title     = {Variational Analysis in Sobolev and BV Spaces: Applications to PDEs and Optimization},
	year      = {2014},
	publisher = {Society for Industrial and Applied Mathematics},
	address   = {Philadelphia, PA}
}

@article{chen2025uncertainty,
	title={Uncertainty-Guided Adaptive Correction for Semi-Supervised Medical Image Segmentation},
	author={Chen, Xi and Tong, Lyuyang and Zhao, Huangxuan and Du, Bo},
	journal={IEEE Transactions on Image Processing},
	volume={34},
	pages={7975--7988},
	year={2025},
	publisher={IEEE}
}

@article{song2025,
	Author = {Song, Fanghui and Sun, Jiebao and Shi, Shengzhu and Guo, Zhichang and
	Wu, Boying},
	Title = {An Effective Level Set Method With Molecular Beam Epitaxy Regularization
	for Color-Texture Image Segmentation},
	Journal = {STUDIES IN APPLIED MATHEMATICS},
	Year = {2025},
	Volume = {155},
	Number = {4},
	Month = {OCT},
	DOI = {10.1111/sapm.70128},
	Article-Number = {e70128},
	ISSN = {0022-2526},
	EISSN = {1467-9590},
	ResearcherID-Numbers = {Song, Fanghui/LPR-1800-2024},
	Unique-ID = {WOS:001605966000009},
}

@article{li2008minimization,
  title={Minimization of region-scalable fitting energy for image segmentation},
  author={Li, Chunming and Kao, Chiu-Yen and Gore, John C and Ding, Zhaohua},
  journal={IEEE transactions on image processing},
  volume={17},
  number={10},
  pages={1940--1949},
  year={2008},
  publisher={IEEE}
}

\end{document}